\pdfoutput=1 

\documentclass[11pt,onecolumn,draftcls,journal]{IEEEtran}
\newif\iffinal
\finalfalse

\ifCLASSINFOpdf
  \usepackage[pdftex]{graphicx}
  \DeclareGraphicsExtensions{.pdf}
\else
\fi
\usepackage[utf8]{inputenc}
\usepackage{flushend}
\usepackage{enumerate}

\ifCLASSOPTIONcompsoc
  \usepackage[caption=false,font=normalsize,labelfont=sf,textfont=sf]{subfig}
\else
  \usepackage[caption=false,font=footnotesize]{subfig}
\fi

\usepackage{url}
\usepackage{hyperref}
\usepackage{balance}

\usepackage{multirow} 
\usepackage{marvosym} 

\usepackage[table]{xcolor}

\begin{document}
\sloppy

\noindent \copyright 2015 IEEE. Personal use of this material is permitted. Permission from IEEE must be obtained for all other uses, in any current or future media, including reprinting/republishing this material for advertising or promotional purposes, creating new collective works, for resale or redistribution to servers or lists, or reuse of any copyrighted component of this work in other works.
\\

\noindent Pre-print of article that will appear in IEEE TRANSACTIONS ON INFORMATION FORENSICS AND SECURITY (T.IFS) Special Issue on Biometric Spoofing and Countermeasures

\newpage

\title{Deep Representations for Iris, Face, and Fingerprint Spoofing Detection}
\author{David~Menotti\textsuperscript{\Yinyang},~\IEEEmembership{Member,~IEEE,}
        Giovani~Chiachia\textsuperscript{\Yinyang},
        Allan~Pinto,~\IEEEmembership{Student Member,~IEEE,}
        William~Robson~Schwartz,~\IEEEmembership{Member,~IEEE,}
        Helio~Pedrini,~\IEEEmembership{Member,~IEEE,}\\
        Alexandre~Xavier~Falc\~{a}o,~\IEEEmembership{Member,~IEEE,}
        and~Anderson~Rocha,~\IEEEmembership{Member,~IEEE}
\thanks{Copyright (c) 2015 IEEE. Personal use of this material is permitted. However, permission to use this material for any other purposes must be obtained from the IEEE by sending a request to pubs-permissions@ieee.org}
\thanks{D. Menotti, G. Chiachia, A. Pinto, H. Pedrini, A. X. Falc\~{a}o and A. Rocha are with the Institute of Computing (IC), University of Campinas, Campinas (UNICAMP), SP, 13083-852, Brazil. email: menottid@gmail.com, \{chiachia,allan.pinto,helio.pedrini,afalcao,anderson.rocha\}@ic.unicamp.br.}
\thanks{W. R. Schwartz is with the Computer Science Department (DCC), Federal University of Minas Gerais (UFMG), Belo Horizonte, MG, 31270-010, Brazil. email: william@dcc.ufmg.br.}
\thanks{D. Menotti is also with Computing Department (DECOM), Federal University of Ouro Preto (UFOP), Ouro Preto, MG, 35400-000, Brazil (he has spent his sabbatical year (2013-2014) at IC-UNICAMP). email: menotti@iceb.ufop.br.}
\thanks{{\small \Yinyang} These authors contributed equally to this work.}%
\thanks{Manuscript received July 3rd, 2014; revised November 18th, 2014; accepted January 27th, 2015.}%
}


\maketitle

\begin{abstract}
Biometrics systems have significantly improved person
identification and authentication, playing an important role in
personal, national, and global security. However, these systems
might be deceived (or ``spoofed'') and, despite the recent advances
in spoofing detection, current solutions often rely on domain
knowledge, specific biometric reading systems, and attack types. We
assume a very limited knowledge about biometric spoofing at the
sensor to derive outstanding spoofing detection systems for iris,
face, and fingerprint modalities based on two deep learning
approaches.
The first approach consists of learning suitable convolutional 
network architectures for each domain, while the second approach
focuses on learning the weights of the network via back-propagation. We
consider nine biometric spoofing benchmarks --- each one containing
real and fake samples of a given biometric modality and attack type
--- and learn deep representations for each benchmark by combining and
contrasting the two learning approaches. This strategy not only
provides better comprehension of how these approaches interplay, but
also creates systems that exceed the best known results in eight out
of the nine benchmarks. The results strongly indicate that spoofing
detection systems based on convolutional networks can be robust to
attacks already known and possibly adapted, with little effort, to
image-based attacks that are yet to come.

\end{abstract}


%
\IEEEpeerreviewmaketitle

\section{Introduction}

\IEEEPARstart{B}{iometrics} human characteristics and traits can successfully allow people identification and authentication and have been widely used for access control, surveillance, and also in national and global security systems~\cite{Jain:2008}.
In the last few years, due to the recent technological improvements for data acquisition, storage and processing, and also the scientific advances in computer vision, pattern recognition, and machine learning, several biometric modalities have been largely applied to person recognition, ranging from traditional fingerprint to face, to iris, and, more recently, to vein and blood flow. Simultaneously, various \emph{spoofing attacks} techniques have been created to defeat such biometric systems. 

There are several ways to spoof a biometric system~\cite{Rathgeb:ICPR:2010,Rathgeb:CVPRW:2011}. Indeed, previous studies show at least eight different points of attack~\cite{Galbally:DATABASE:2007,Ratha:AVBPA:2001} that can be divided into two main groups: \emph{direct} and \emph{indirect} attacks. The former considers the possibility to generate synthetic biometric samples, and is the first vulnerability point of a biometric security system acting at the sensor level. The latter includes all the remaining seven points of attacks and requires different levels of knowledge about the system, e.g., the matching algorithm used, the specific feature extraction procedure, database access for manipulation, and also possible weak links in the communication channels within the system.


Given that the most vulnerable part of a system is its acquisition sensor, attackers have mainly focused on direct spoofing. This is possibly because a number of biometric traits can be easily forged with the use of common apparatus and consumer electronics to imitate real biometric readings (e.g., stampers, printers, displays, audio recorders). 
In response to that, several biometric spoofing benchmarks have been recently proposed, allowing researchers to make steady progress in the conception of anti-spoofing systems. Three relevant modalities in which spoofing detection has been investigated are iris, face, and fingerprint. Benchmarks across these modalities usually share the common characteristic of being image- or video-based.



In the context of irises, attacks are normally performed using printed iris images~\cite{Sequeira:IJCB:2014} or, more interestingly, cosmetic contact lenses~\cite{Bowyer:Computer:2014,Yadav:TIFS:2014}.
With faces, impostors can present to the acquisition sensor a photography, a digital video~\cite{Chingovska:BIOSEG:2012}, or even a 3D mask~\cite{Erdogmus:BTAS:2013} of a valid user. 
%
%
For fingerprints, the most common spoofing method consists of using artificial replicas~\cite{Ghiani:ICB:2013} created in a cooperative way, where a mold of the fingerprint is acquired with the cooperation of a valid user and is used to replicate the user's fingerprint with different materials, including gelatin, latex, play-doh or silicone.

The success of an anti-spoofing method is usually connected to the modality for which it was designed. In fact, such systems often rely on expert knowledge to engineer features that are able to capture acquisition telltales left by specific types of attacks. However, the need of custom-tailored solutions for the myriad possible attacks might be a limiting constraint. Small changes in the attack could require the redesign of the entire system.

In this paper, we do not focus on custom-tailored solutions. Instead, inspired by the recent success of Deep Learning in several vision tasks~\cite{Ciresan:2010,Krizhevsky:2012,Ciresan:2012,Ouyang:2014,Taigman:2014}, and by the ability of the technique to leverage data, we focus on two general-purpose approaches to build image-based anti-spoofing systems with convolutional networks for several attack types in three biometric modalities, namely iris, face, and fingerprint. The first technique that we explore is hyperparameter optimization of network architectures~\cite{Pinto:2009,Bergstra:2012} that we henceforth call \emph{architecture optimization}, while the second lies at the core of convolutional networks and consists of learning filter weights via the well-known back-propagation~\cite{LeCun:1998} algorithm, hereinafter referred to as \emph{filter optimization}.

\begin{figure}
\begin{center}
\includegraphics[width=0.9\linewidth]{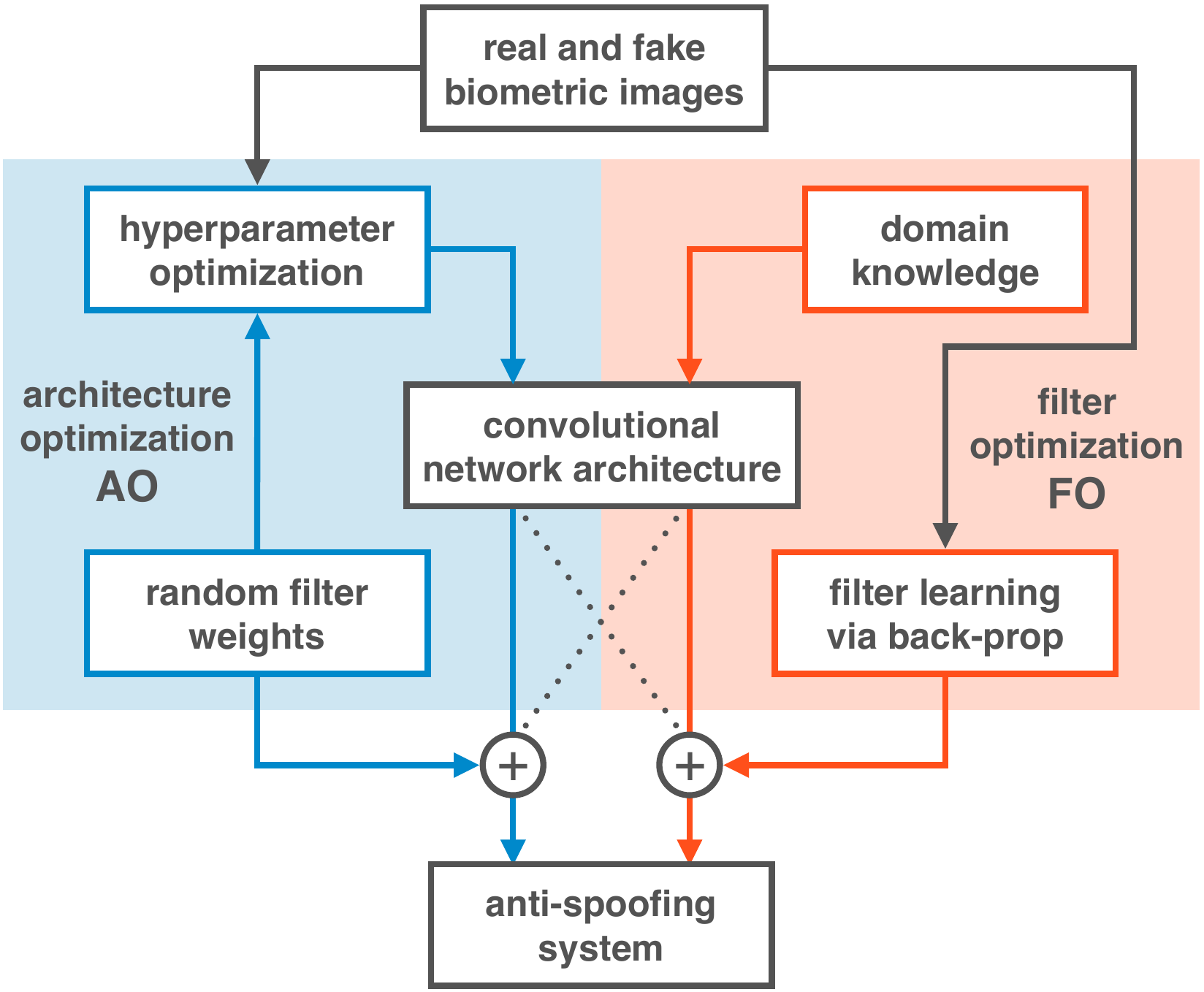} 
\end{center}
\caption{Schematic diagram detailing how anti-spoofing systems are
  built from spoofing detection benchmarks. Architecture optimization
  (AO) is shown on the left and filter optimization (FO) on the
  right. In this work, we not only evaluate AO and FO in separate, but
  also in combination, as indicated by the crossing dotted lines.}
\label{f.schema}
\end{figure}

Fig.~\ref{f.schema} illustrates how such techniques are used. The architecture optimization (AO) approach is presented on the left and is highlighted in blue while the filter optimization (FO) approach is presented on the right and is highlighted in red. As we can see, AO is used to search for good architectures of convolutional networks in a given spoofing detection problem and uses convolutional filters whose weights are set at random in order to make the optimization practical. This approach assumes little a priori knowledge about the problem, and is an area of research in deep learning that has been successful in showing that the architecture of convolutional networks, by themselves, is of extreme importance to performance~\cite{Pinto:2009,Bergstra:2012,Saxe:2011,Pinto:2011b,Bergstra:2011,Bergstra:2013}. In fact, the only knowledge AO assumes about the problem is that it is approachable from a computer vision perspective.

Still in Fig~\ref{f.schema}, FO is carried out with back-propagation in a predefined network architecture. This is a longstanding approach for building convolutional networks that has recently enabled significant strides in computer vision, specially because of an understanding of the learning process, and the availability of plenty of data and processing power~\cite{Krizhevsky:2012,Taigman:2014,Simonyan:2014}. Network architecture in this context is usually determined by previous knowledge of related problems.


In general, we expect AO to adapt the architecture to the problem in hand and FO to model important stimuli for discriminating fake and real biometric samples. 
We evaluate AO and FO not only in separate, but also in combination, i.e., architectures learned with AO are used for FO as well as previously known good performing architectures are used with random filters.
This explains the crossing dotted lines in the design flow of Fig~\ref{f.schema}.

As our experiments show, the benefits of evaluating AO and FO apart and later combining them to build anti-spoofing systems are twofold.
First, it enables us to have a better comprehension of the interplay between these approaches, something that has been largely underexplored in the literature of convolutional networks. Second, it allows us to build systems with outstanding performance in all nine publicly available benchmarks considered in this work.

The first three of such benchmarks consist of spoofing attempts for iris recognition systems, Biosec~\cite{Ruiz-Albacete:BIOID:2008}, Warsaw~\cite{Czajka:MMAR:2013}, and MobBIOfake~\cite{Sequeira:VISAPP:2014:base}.
Replay-Attack~\cite{Chingovska:BIOSEG:2012} and 3DMAD~\cite{Erdogmus:BTAS:2013} are the benchmarks considered for faces, while Biometrika, CrossMatch, Italdata, and Swipe are the fingerprint benchmarks here considered, all them recently used in the 2013 Fingerprint Liveness Detection Competition (LivDet'13)~\cite{Ghiani:ICB:2013}.

Results outperform state-of-the-art counterparts in eight of the nine cases and observe a balance in terms of performance between AO and FO, with one performing better than the other depending on the sample size and problem difficulty. In some cases, we also show that when both approaches are combined, we can obtain performance levels that neither one can obtain by itself. Moreover, by observing the behaviour of AO and FO, we take advantage of domain knowledge to propose a single new convolutional architecture that push performance in five problems even further, sometimes by a large margin, as in CrossMatch (68.80\% \emph{v.} 98.23\%).  

The experimental results strongly indicate that convolutional networks can be readily used for robust spoofing detection.  
Indeed, we believe that data-driven solutions based on deep representations might be a valuable direction to this field of research, allowing the construction of systems with little effort even to image-based attack types yet to come.

We organized the remainder of this work into five sections.
Section~\ref{sec:relatedwork} presents previous anti-spoofing systems for the three biometric modalities covered in this paper, while Section~\ref{sec:databases} presents the considered benchmarks. Section~\ref{sec:methodology} describes the methodology adopted for architecture optimization (AO) and filter optimization (FO) while Section~\ref{sec:experiments} presents experiments, results, and comparisons with state-of-the-art methods. Finally, Section~\ref{sec:conclusions} concludes the paper and discusses some possible future directions.

\section{Related Work}
\label{sec:relatedwork}
In this section, we review anti-spoofing related work for iris, face, and fingerprints, our focus in this paper.

\subsection{Iris Spoofing}
Daugman~\cite[Section 8 -- Countermeasures against Subterfuge]{Daugman:1999}\footnote{It also appears in a lecture of Daugman at IBC 2004~\cite{ Daugman:IBC:2004}.} was one of the first authors to discuss the feasibility of some attacks on iris recognition systems. The author proposed the use of Fast Fourier Transform to verify the high frequency spectral magnitude in the frequency domain.


The solutions for iris liveness detection available in the literature range from active solutions relying on  special acquisition hardware~\cite{Lee:LNCS:2005,Pacut:2006,Kanematsu:2007} to software-based solutions relying on texture analysis of the  effects of an attacker using color contact lenses with someone else's pattern printed onto them~\cite{Wei:2008}. Software-based solutions have also explored the effects of cosmetic contact lenses~\cite{Kohli:ICB:2013,Doyle:BTAS:2013,Bowyer:Computer:2014,Yadav:TIFS:2014}; pupil constriction~\cite{Huang:WACV:2013}; and multi biometrics of  electroencephalogram (EEG) and iris together~\cite{Kathikeyan:ICCCA:2012}, among others.

Galbally et al.~\cite{Galbally:ICB:2012} investigated 22 image quality measures (e.g., focus, motion, occlusion, and pupil dilation).
The best features are selected through sequential floating feature selection (SFFS)~\cite{Pudil:PRL:1994} to feed a quadratic discriminant classifier. The authors validated the work on the BioSec~\cite{Fierrez-Aguilar:2007,Ruiz-Albacete:BIOID:2008} benchmark. Sequeira et al.~\cite{Sequeira:VISAPP:2014} 
also explored image quality measures~\cite{Galbally:ICB:2012} and three classification techniques validating the work on the BioSec~\cite{Fierrez-Aguilar:2007,Ruiz-Albacete:BIOID:2008} and Clarkson~\cite{LivDet:Iris:2013} benchmarks and introducing the MobBIOfake benchmark comprising 800 iris images from the MobBIO multimodal database~\cite{Sequeira:VISAPP:2014:base}.

Sequeira et al.~\cite{Sequeira:IJCNN:2014} extended upon previous works also exploring quality measures. They first used a feature selection step on the features of the studied methods to obtain the ``best features'' and then used well-known classifiers for the decision-making. In addition, they applied iris segmentation~\cite{Monteiro:CCIS:2014} to obtaining the iris contour and adapted the feature extraction processes to the resulting non-circular iris regions. The validation considered five datasets (BioSec~\cite{Fierrez-Aguilar:2007,Ruiz-Albacete:BIOID:2008}, MobBIOfake~\cite{Sequeira:VISAPP:2014:base}, Warsaw~\cite{Czajka:MMAR:2013}, Clarkson~\cite{LivDet:Iris:2013} and NotreDame~\cite{Doyle:2014:base}.

Textures have also been explored for iris liveness detection. In the recent MobILive\footnote{MobLive 2014, Intl. Joint Conference on Biometrics (IJCB).}~\cite{Sequeira:IJCB:2014} iris spoofing detection competition, the winning team explored three texture descriptors: Local Phase Quantization (LPQ)~\cite{Ojansivu:ISP:2008}, Binary Gabor Pattern~\cite{Zhang:ICIP:2012}, and Local Binary Pattern (LBP)~\cite{Ojala:TPAMI_2002}.

Analyzing printing regularities left in printed irises, Czajka~\cite{Czajka:MMAR:2013} explored some peaks in the frequency spectrum were associated to spoofing attacks. For validation, the authors introduced the Warsaw dataset containing 729 fake images and 1,274 images of real eyes. 
In~\cite{LivDet:Iris:2013}, The First Intl. Iris Liveness Competition in 2013, the Warsaw database was also evaluated, however, the best reported result achieved $11.95\%$ of FRR and $5.25\%$ of FAR by the University of Porto team.



Sun et al.~\cite{Sun:TPAMI:2014} recently proposed a general framework for iris image classification based on a Hierarchical Visual Codebook (HVC). The HVC encodes the texture primitives of iris images and is based on two existing bag-of-words models. The method achieved state-of-the-art performance for iris spoofing detection, among other tasks.


In summary, iris anti-spoofing methods have explored hard-coded features through image-quality metrics, texture patterns, bags-of-visual-words and noise artifacts due to the recapturing process. The performance of such solutions vary significantly from dataset to dataset. Differently, here we propose the automatically extract vision meaningful features directly from the data using deep representations.

\subsection{Face Spoofing}

We can categorize the face anti-spoofing methods into four groups~\cite{Schwartz:IJCB:2011}: user behavior modeling, methods relying on extra devices~\cite{Yi:2014}, methods relying on user cooperation and, finally, data-driven characterization methods. In this section, we review data-driven characterization methods proposed in literature, the focus of our work herein.

M\"{a}\"{a}tt\"{a} et al.~\cite{Maatta:IJCB:2011} used LBP operator for capturing printing artifacts and micro-texture patterns added in the fake biometric samples during acquisition. Schwartz et al.~\cite{Schwartz:IJCB:2011} explored color, texture, and shape of the face region and used them with Partial Least Square (PLS) classifier for deciding whether a biometric sample is fake or not. Both works validated the methods with the Print Attack benchmark~\cite{Anjos:IJCB:2011}. Lee et al.~\cite{Lee:ICASSP:2013} also explored image-based attacks and proposed the frequency entropy analysis for spoofing detection.



Pinto et al.~\cite{Pinto:SIBGRAPI:2012} pioneered research on video-based face spoofing detection. They proposed visual rhythm analysis to capture temporal information on face spoofing attacks.



Mask-based face spoofing attacks have also been considered thus far. Erdogmus et al.~\cite{Erdogmus:BIOSIG:2013} dealt with the problem through Gabor wavelets: local Gabor binary pattern histogram sequences~\cite{Zhang:ICCV:2005} and Gabor graphs~\cite{Wiskott:TPAMI:1997} with a Gabor-phase based similarity measure~\cite{Gunther:ICANN:2012}. Erdogmus \& Marcel~\cite{Erdogmus:BTAS:2013} introduced the 3D Mask Attack database (3DMAD), a public available 3D spoofing database, recorded with Microsoft Kinect sensor.


Kose et al.~\cite{Kose:ICASSP:2013} demonstrated that a face verification system is vulnerable to mask-based attacks and, in another work, Kose et al.~\cite{Kose:FG:2013} evaluated the anti-spoofing method proposed by M\"{a}\"{a}tt\"{a} et al.~\cite{Maatta:IJCB:2011} (originally proposed to detect photo-based spoofing attacks). 
Inspired by the work of Tan et al.~\cite{Tan:ECCV:2010}, Kose et al.~\cite{Kose:DSP:2013} evaluated a solution based on reflectance to detect attacks performed with 3D masks. 


Finally, Pereira et al.~\cite{Pereira:ICB:2013} proposed a score-level fusion strategy in order to detect various types of attacks. 
In a follow-up work, Pereira et al.~\cite{Pereira:JIVP:2014} proposed an anti-spoofing solution based on the dynamic texture, a spatio-temporal version of the original LBP. Results showed that LBP-based dynamic texture description has a higher effectiveness than the original LBP.

In summary, similarly to iris spoofing detection methods, the available solutions in the literature mostly deal with the face spoofing detection problem through texture patterns (e.g., LBP-like detectors), acquisition telltales (noise), and image quality metrics. Here, we approach the proplem by extracting meaningful features directly from the data regardless of the input type (image, video, or 3D masks).

\subsection{Fingerprint Spoofing}
We can categorize fingerprint spoofing detection methods roughly into two groups: hardware-based (exploring extra sensors) and software-based solutions (relying only on the information acquired by the standard acquisition sensor of the authentication system)~\cite{Ghiani:ICB:2013}. 


Galbally et al.~\cite{Galbally:BIDS:2009} proposed a set of feature for fingerprint liveness detection based on quality measures such as ridge strength or directionality, ridge continuity, ridge clarity, and integrity of the ridge-valley structure. The validation considered the three benchmarks used in LivDet 2009 -- Fingerprint competition~\cite{Marcialis:ICIAP:2009} captured with different optical sensors: Biometrika, CrossMatch, and Identix. Later work~\cite{Galbally:FGCS:2012} explored the method in the presence of gummy fingers.



Ghiani et al.~\cite{Ghiani:ICPR:2012} explored LPQ~\cite{Ojansivu:ISP:2008}, a method for representing all spectrum characteristics in a compact feature representation form. The validation considered the four benchmarks used in the LivDet 2011 -- Fingerprint competition~\cite{Yambay:ICB:2012}.



Gragnaniello et al.~\cite{Gragnaniello:BIOMS:2013} explored the Weber Local Image Descriptor (WLD) for liveness detection, well suited to high-contrast patterns such as the ridges and valleys of fingerprints images. In addition, WLD is robust to noise and illumination changes. The validation considered the LivDet 2009 and LivDet 2011 -- Fingerprint competition datasets.


Jia et al.~\cite{Jia:ICB:2013} proposed a liveness detection scheme based on Multi-scale Block Local Ternary Patterns (MBLTP). 
Differently of the LBP, the Local Ternary Pattern operation is done on the average value of the block instead of the pixels being more robust to noise. 
The validation considered the LivDet 2011 -- Fingerprint competition benchmarks.

Ghiani et al.~\cite{Ghiani:BTAS:2013} explored Binarized Statistical Image Features (BSIF) originally proposed by Kannala et al.~\cite{Kannala:ICPR:2012}. The BSIF was inspired in the LBP and LPQ methods. In contrast to LBP and LPQ approaches, BSIF learns a filter set by using statistics of natural images~\cite{Hyvrinen:NIS:2009}. The validation considered the LivDet 2011 -- Fingerprint competition benchmarks.

Recent results reported in the LivDet 2013 Fingerprint Liveness Detection Competition~\cite{Ghiani:BTAS:2013} show that fingerprint spoofing attack detection task is still an open problem with results still far from a perfect classification rate.

We notice that most of the groups approach the problem with hard-coded features sometimes exploring quality metrics related to the modality (e.g., directionality and ridge strength), general texture patterns (e.g., LBP-, MBLTP-, and LPQ-based methods), and filter learning through natural image statistics. This last approach seems to open a new research trend, which seeks to model the problem learning features directly from the data. We follow this approach in this work, assuming little a priori knowledge about acquisition-level biometric spoofing and exploring deep representations of the data.


\subsection{Multi-modalities}

Recently, Galbally et al.~\cite{Galbally:TIP:2014} proposed a general approach based on 25 image quality features to detect spoofing attempts in face, iris, and fingerprint biometric systems. Our work is similar to theirs in goals, but radically different with respect to the methods.
Instead of relying on prescribed image quality features, we build features that would be hardly thought by a human expert with AO and FO.
Moreover, here we evaluate our systems in more recent and updated benchmarks.


\section{Benchmarks}
\label{sec:databases}

In this section, we describe the benchmarks (datasets) that we consider in this work. All them are publicly available upon request and suitable for evaluating countermeasure methods to iris, face and fingerprint spoofing attacks. Table~\ref{tab:databases} shows the major features of each one and in the following we describe their details.

\begin{table*}[tb!]
\begin{center}
\caption{Main features of the benchmarks considered herein.}
\label{tab:databases}
\iffinal
\else
\tiny
\fi
\begin{tabular}{clccrrrcrrrcrrr}
\hline
\multirow{2}{*}{Modality}
& \multirow{2}{*}{Benchmark/Dataset}
                                            & \multirow{2}{*}{Color}
                                                    &      Dimension
                                                                         &\multicolumn{3}{c}{\# Training}
                                                                                               && \multicolumn{3}{c}{\# Testing} 
                                                                                                                     && \multicolumn{3}{c}{\# Development} \\
\cline{5-7}\cline{9-11}\cline{13-15}
&                                            &       & $cols \times rows$ & Live & Fake & Total && Live & Fake & Total && Live & Fake & Total \\
\hline
\hline
\multirow{3}{*}{Iris}
&Warsaw~\cite{Czajka:MMAR:2013}              & No    & $640 \times  480$ &  228 &  203 &   431 &&  624 &  612 &  1236 \\
&Biosec~\cite{Ruiz-Albacete:BIOID:2008}      & No    & $640 \times  480$ &  200 &  200 &   400 &&  600 &  600 &  1200 \\
&MobBIOfake~\cite{Sequeira:VISAPP:2014:base} & Yes   & $250 \times  200$ &  400 &  400 &   800 &&  400 &  400 &   800 \\
\hline
\multirow{2}{*}{Face}
& Replay-Attack~\cite{Chakka:IJCB:2011}      & Yes   & $320 \times  240$ &  600 & 3000 &  3600 && 4000 &  800 &  4800 &&  600 & 3000 & 3600 \\
& 3dMad~\cite{Chingovska:ICB:2013}           & Yes   & $640 \times  480$ &  350 &  350 &   700 &&  250 &  250 &   500 &&  250 &  250 &  500 \\
\hline
\multirow{4}{*}{Fingerprint} 
&Biometrika~\cite{Ghiani:ICB:2013}           & No    & $312 \times  372$ & 1000 & 1000 &  2000 && 1000 & 1000 &  2000 \\
&CrossMatch~\cite{Ghiani:ICB:2013}           & No    & $800 \times  750$ & 1250 & 1000 &  2250 && 1250 & 1000 &  2250 \\
&Italdata~\cite{Ghiani:ICB:2013}             & No    & $640 \times  480$ & 1000 & 1000 &  2000 && 1200 & 1000 &  2000 \\
&Swipe~\cite{Ghiani:ICB:2013}                & No    & $208 \times 1500$ & 1221 &  979 &  2200 && 1153 & 1000 &  2153 \\
\hline

\end{tabular}
\end{center}
\end{table*}

\subsection{Iris Spoofing Benchmarks}

\subsubsection{Biosec} 
This benchmark was created using iris images from $50$ users of the BioSec~\cite{Ruiz-Albacete:BIOID:2008}.
In total, there are $16$ images for each user ($2$ sessions $\times$ $2$ eyes $\times$ $4$ images), totalizing $800$ valid access images. 
To create spoofing attempts, the original images from Biosec were preprocessed to improve quality and printed using an HP Deskjet 970cxi and an HP LaserJet 4200L printers. 
Finally, the iris images were recaptured with the same iris camera used to capture the original images.

\subsubsection{Warsaw} 
This benchmark contains $1274$ images of $237$ volunteers representing valid accesses and $729$ printout images representing spoofing attempts, which were generated by using two printers: (1) a HP LaserJet 1320 used to produce $314$ fake images with $600$ dpi resolution, and (2) a Lexmark C534DN used to produce $415$ fake images with $1200$ dpi resolution. Both real and fake images were captured by an IrisGuard AD100 biometric device.

\subsubsection{MobBIOfake} 
This benchmark contains live iris images and fake printed iris images captured with the same acquisition sensor, i.e., a mobile phone. To generate fake images, the authors first performed a preprocessing in the original images to enhance the contrast. The preprocessed images were then printed with a professional printer on high quality photographic paper.

\subsection{Video-based Face Spoofing Benchmarks}

\subsubsection{Replay-Attack} 
This benchmark contains short video recordings of both valid accesses and video-based attacks of $50$ different subjects. 
To generate valid access videos, each person was recorded in two sessions in a controlled and in an adverse environment with a regular webcam.
Then, spoofing attempts were generated using three techniques:
(1)~\emph{print attack}, which presents to the acquisition sensor hard copies of high-resolution digital photographs printed with a Triumph-Adler DCC 2520 color laser printer; 
(2) \emph{mobile attack}, which presents to the acquisition sensor photos and videos taken with an iPhone using the iPhone screen; 
and (3) \emph{high-definition attack}, in which high resolution photos and videos taken with an iPad are presented to the acquisition sensor using the iPad screen.

\subsubsection{3DMAD} 
This benchmark consists of real videos and fake videos made with people wearing masks. A total of $17$ different subjects were recorded with a Microsoft Kinect sensor, and videos were collected in three sessions. For each session and each person, five videos of $10$ seconds were captured. The 3D masks were produced by \url{ThatsMyFace.com} using one frontal and two profile images of each subject.
All videos were recorded by the same acquisition sensor.

\subsection{Fingerprint Spoofing Benchmarks}

\subsubsection{LivDet2013} 
This dataset contains four sets of real and fake fingerprint readings performed in four acquisition sensors: Biometrika FX2000, Italdata ET10, Crossmatch L Scan Guardian, and Swipe.
For a more realistic scenario, fake samples in Biometrika and Italdata were generated without user cooperation, while fake samples in Crossmatch and Swipe were generated with user cooperation. 
Several materials for creating the artificial fingerprints were used, including gelatin, silicone, latex, among others. 


\subsection{Remark}


Images found in these benchmarks can be observed in Fig.~\ref{fig:datasets} of Section~\ref{sec:experiments}. 
As we can see, variability exists not only across modalities, but also within modalities. Moreover, it is rather unclear what features might discriminate real from spoofed images, which suggests that the use of a methodology able to use data to its maximum advantage might be a promising idea to tackle such set of problems in a principled way.

\section{Methodology}
\label{sec:methodology}

In this section, we present the methodology for architecture optimization (AO) and filter optimization (FO) as well as details about how benchmark images are preprocessed, how AO and FO are evaluated across the benchmarks, and how these methods are implemented.

\subsection{Architecture Optimization (AO)}
\label{sec:ao}

Our approach for AO builds upon the work of Pinto et al.~\cite{Pinto:2009} and Bergstra et al.~\cite{Bergstra:2013}, i.e., fundamental, feedforward convolutional operations are stacked by means of hyperparameter optimization, leading to effective yet simple convolutional networks that do not require expensive filter optimization and from which prediction is done by linear support vector machines (SVMs).

Operations in convolutional networks can be viewed as linear and non-linear transformations that, when stacked, extract high level representations of the input. Here we use a well-known set of operations called (i) \emph{convolution} with a bank of filters, (ii) rectified linear \emph{activation}, (iii) spatial \emph{pooling}, and (iv) \emph{local normalization}. Appendix~\ref{sec:convnet_ops} provides a detailed definition of these operations.

\begin{figure}
\begin{center}
 \includegraphics[width=1.0\linewidth]{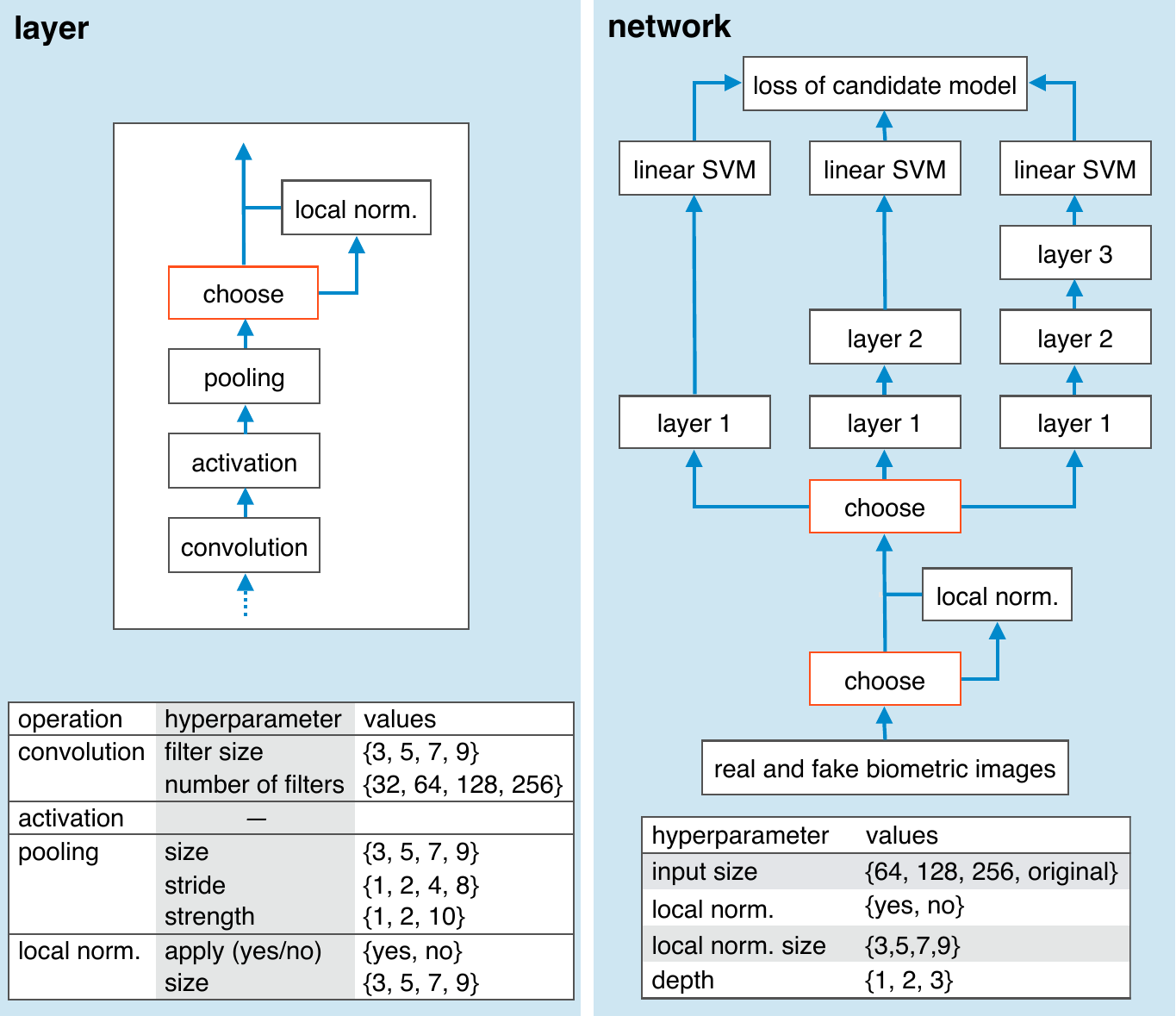}
 \caption{Schematic diagram for architecture optimization (AO) illustrating how operations are stacked in a layer (left) and how the network is instantiated and evaluated according to possible hyperparameter values (right). Note that a three-layered convolutional network of this type has a total of 25 hyperparameters governing both its architecture and its overall behaviour through a particular instance of stacked operations.}
 \label{fig:framework}
\end{center}
\end{figure}

We denote as \emph{layer} the combination of these four operations in the order that they appear in the left panel of Fig.~\ref{fig:framework}. Local normalization is optional and its use is governed by an additional ``yes/no'' hyperparameter. In fact, there are other six hyperparameters, each of a particular operation, that have to be defined in order to instantiate a layer. They are presented in the lower part of the left panel in Fig.~\ref{fig:framework} and are in accordance to the definitions of Appendix~\ref{sec:convnet_ops}.

Considering one layer and possible values of each hyperparameter, there are over 3,000 possible layer architectures, and this number grows exponentially with the number of layers, which goes up to three in our case (Fig.~\ref{fig:framework} right panel). In addition, there are network-level hyperparameters, such as the size of the input image, that expand possibilities to a myriad potential architectures.
 
The overall set of possible hyperparameter values is called \emph{search space}, which in this case is discrete and contains variables that are only meaningful in combination with others. For example, hyperparameters of a given layer are just meaningful if the candidate architecture has actually that number of layers.
In spite of the intrinsic difficulty in optimizing architectures in this space, \emph{random search} has played and important role in problems of this type~\cite{Pinto:2009,Bergstra:2012} and it is the strategy of our choice due to its effectiveness and simplicity.

We can see in Fig.~\ref{fig:framework} that a three-layered network has a total of 25 hyperparameters, seven per layer and four at network level. They are all defined in Appendix~\ref{sec:convnet_ops} with the exception of \emph{input size}, which seeks to determine the best size of the image's greatest axis (rows or columns) while keeping its aspect ratio. Concretely, random search in this paper can be described as follows:

\begin{enumerate}
\item Randomly --- and uniformly, in our case --- sample values from the hyperparameter \emph{search space};
\item Extract features from real and fake training images with the candidate architecture;
\item Evaluate the architecture according to an \emph{optimization objective} based on linear SVM scores;
\item Repeat steps 1--3 until a \emph{termination criterion} is met;
\item Return the best found convolutional architecture.
\end{enumerate}

Even though there are billions of possible networks in the search space (Fig.~\ref{fig:framework}), it is important to remark that not all candidate networks are valid. For example, a large number of candidate architectures (i.e., points in the search space) would produce representations with spatial resolution smaller than one pixel. Hence, they are naturally unfeasible.
Additionally, in order to avoid very large representations, we discard in advance candidate architectures whose intermediate layers produce representations of over 600K elements or whose output representation has over 30K elements.

Filter weights are randomly generated for AO. This strategy has been successfully used in the vision literature~\cite{Pinto:2009,Saxe:2011,Pinto:2011b,Jarrett:2009} and is essential to make AO practical, avoiding the expensive filter optimization (FO) part in the evaluation of candidate architectures. We sample weights from a uniform distribution $U(0,1)$ and normalize the filters to zero mean and unit norm in order to ensure that they are spread over the unit sphere. When coupled with rectified linear activation (Appendix~\ref{sec:convnet_ops}), this sampling enforces sparsity in the network by discarding about $50$\% of the expected filter responses, thereby improving the overall robustness of the feature extraction.

A candidate architecture is evaluated by first extracting deep representations from real and fake images and later training hard-margin linear SVMs ($C$=$10^5$) on these representations.
We observed that the sensitivity of the performance measure was saturating with traditional 10-fold cross validation (CV) in some benchmarks. Therefore, we opted for a different validation strategy. Instead of training on nine folds and validating on one, we train on one fold and validate on nine. Precisely, the \emph{optimization objective} is the mean detection accuracy obtained from this adapted cross-validation scheme, which is maximized during the optimization.

For generating the 10 folds, we took special care in putting all samples of an individual in the same fold to enforce robustness to cross-individual spoofing detection in the optimized architectures.
Moreover, in benchmarks where we have more than one attack type (e.g., Replay-Attack and LivDet2013, see Section~\ref{sec:databases}), we evenly distributed samples of each attack type across all folds in order to enforce that candidate architectures are also robust to different types of attack.

Finally, the \emph{termination criterion} of our AO procedure simply consists of counting the number of valid candidate architectures and stopping the optimization when this number reaches 2,000.

\subsection{Filter Optimization (FO)}
\label{sec:fo}

\begin{figure}
\begin{center}
\includegraphics[width=1.0\linewidth]{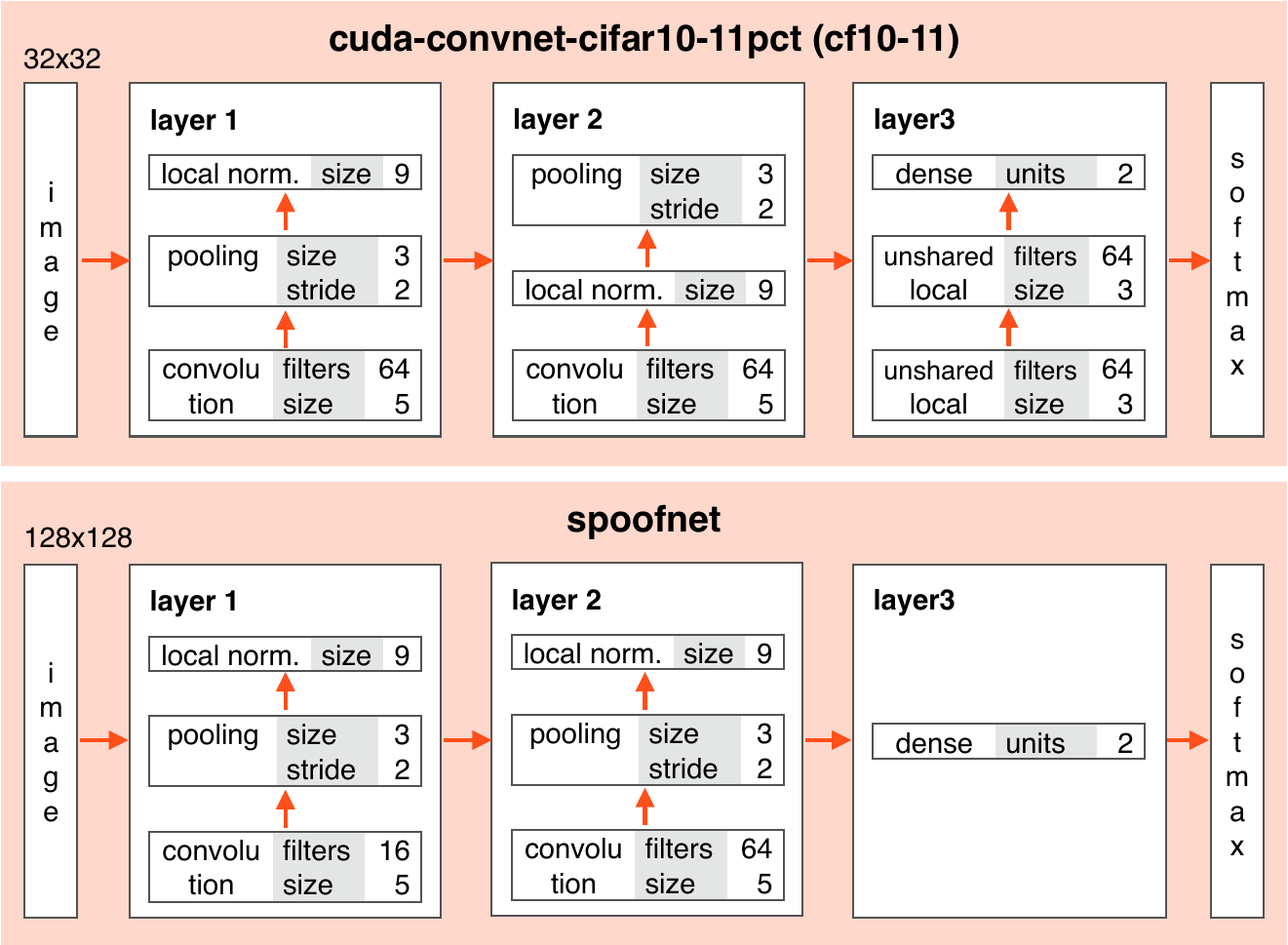}
\caption{
Architecture of convolutional network found in the Cuda-convnet library~\cite{Krizhevsky:cuda-convnet:2012} and here used as reference for filter optimization (\emph{cf10-11}, top). Proposed network architecture extending upon~\emph{cf10-11} to better suiting spoofing detection problems (\emph{spoofnet}, bottom). Both architectures are typical examples where domain knowledge has been incorporated for increased performance.
}
\label{fig:cudaconvnet}
\end{center}
\end{figure}

We now turn our attention to a different approach for tackling the problem. Instead of optimizing the architecture, we explore the filter weights and how to learn them for better characterizing real and fake samples. Our approach for FO is at the origins of convolutional networks and consists of learning filter weights via the well-known back-propagation algorithm~\cite{LeCun:1998}. Indeed, due to a refined understanding of the optimization process and the availability of plenty of data and processing power, back-propagation has been the gold standard method in deep networks for computer vision in the last years~\cite{Krizhevsky:2012,Simonyan:2014,Zeiler:2014}.

For optimizing filters, we need to have an already defined architecture. We start optimizing filters with a standard public convolutional network and training procedure. This network is available in the Cuda-convnet library~\cite{Krizhevsky:cuda-convnet:2012} and is currently one of the best performing architectures in CIFAR-10,\footnote{\url{http://www.cs.toronto.edu/~kriz/cifar.html}} a popular computer vision benchmark in which such network achieves 11\% of classification error. Hereinafter, we call this network \emph{cuda-convnet-cifar10-11pct}, or simply \emph{cf10-11}.

Fig.~\ref{fig:cudaconvnet} depicts the architecture of \emph{cf10-11} in the top panel and is a typical example where domain knowledge has been incorporated for increased performance. We can see it as a three-layered network in which the first two layers are convolutional, with operations similar to the operations used in architecture optimization (AO). In the third layer, \emph{cf10-11} has two sublayers of unshared local filtering and a final fully-connected sublayer on top of which softmax regression is performed. A detailed explanation of the operations in \emph{cf10-11} can be found in~\cite{Krizhevsky:cuda-convnet:2012}.

In order to train~\emph{cf10-11} in a given benchmark, we split the training images into four batches observing the same balance of real and fake images. After that, we follow a procedure similar to the original\footnote{\url{https://code.google.com/p/cuda-convnet/wiki/Methodology}.} for training~\emph{cf10-11} in all benchmarks, which can be described as follows:

\begin{enumerate}
\item For 100 epochs, train the network with a learning rate of $10^{-3}$ by considering the first three batches for training and the fourth batch for validation;
\item For another 40 epochs, resume training now considering all four batches for training;
\item Reduce the learning rate by a factor of 10, and train the network for another 10 epochs;
\item Reduce the learning rate by another factor of 10, and train the network for another 10 epochs.
\end{enumerate}

After evaluating filter learning on the \emph{cf10-11} architecture, we also wondered how filter learning could benefit from an optimized architecture incorporating domain-knowledge of the problem. Therefore, extending upon the knowledge obtained with AO as well as with training~\emph{cf10-11} in the benchmarks, we derived a new architecture for spoofing detection that we call~\emph{spoofnet}. Fig.~\ref{fig:cudaconvnet} illustrates this architecture in the bottom panel and has three key differences as compared to~\emph{cf10-11}. First, it has 16 filters in the first layer instead of 64. Second, operations in the second layer are stacked in the same order that we used when optimizing architectures (AO). Third, we removed the two unshared local filtering operations in the third layer, as they seem inappropriate in a problem where object structure is irrelevant.

These three modifications considerably dropped the number of weights in the network and this, in turn, allowed us to increase of size of the input images from $32\times32$ to $128\times128$. This is the fourth and last modification in~\emph{spoofnet}, and we believe that it might enable the network to be more sensitive to subtle local patterns in the images.

In order to train~\emph{spoofnet}, the same procedure used to train~\emph{cf10-11} is considered except for the initial learning rate, which is made $10^{-4}$, and for the number of epochs in each step, which is doubled. These modifications were made because of the decreased learning capacity of the network.

Finally, in order to reduce overfitting, data augmentation is used for training both networks according to the procedure of~\cite{Krizhevsky:2012}. For \emph{cf10-11}, five $24\times24$ image patches are cropped out from the $32\times32$ input images. These patches correspond to the four corners and central region of the original image, and their horizontal reflections are also considered. Therefore, ten training samples are generated from a single image. For \emph{spoofnet}, the procedure is the same except for the fact that input images have $128\times128$ pixels and cropped regions are of $112\times112$ pixels. During prediction, just the central region of the test image is considered.

\subsection{Elementary Preprocessing}
\label{sec:preproc}

A few basic preprocessing operations were executed on face and fingerprint images in order to properly learn representations for these benchmarks. 
This preprocessing led to images with sizes as presented in Table~\ref{tab:databases:experiments} and are described in the next two sections.

\begin{table}[tb!]
\begin{center}
\caption{Input image dimensionality after basic preprocessing on face and fingerprint images (highlighted). See Section~\ref{sec:preproc} for details.}
\label{tab:databases:experiments}
\begin{tabular}{clc}
\hline
\multirow{2}{*}{Modality}
& \multirow{2}{*}{Benchmark}
                                             & Dimensions \\
&                                            & $columns \times rows$ \\
\hline
\hline
\multirow{3}{*}{Iris}
&Warsaw~\cite{Czajka:MMAR:2013}              & $640 \times  480$ \\
&Biosec~\cite{Ruiz-Albacete:BIOID:2008}      & $640 \times  480$ \\
&MobBIOfake~\cite{Sequeira:VISAPP:2014:base} & $250 \times  200$ \\
\hline
\multirow{2}{*}{Face}
& Replay-Attack~\cite{Chakka:IJCB:2011}      & $\mathbf{200 \times  200}$ \\
& 3DMAD~\cite{Chingovska:ICB:2013}           & $\mathbf{200 \times  200}$ \\
\hline
\multirow{4}{*}{Fingerprint} 
&Biometrika~\cite{Ghiani:ICB:2013}           & $\mathbf{312 \times  372}$ \\
&CrossMatch~\cite{Ghiani:ICB:2013}           & $\mathbf{480 \times  675}$ \\
&Italdata~\cite{Ghiani:ICB:2013}             & $\mathbf{384 \times  432}$ \\
&Swipe~\cite{Ghiani:ICB:2013}                & $\mathbf{187 \times  962}$ \\
\hline

\end{tabular}
\end{center}
\end{table}

\subsubsection{Face Images}

Given that the face benchmarks considered in this work are video-based, we first evenly subsample 10 frames from each input video. Then, we detect the face position using Viola \& Jones~\cite{Viola:IJCV:2001} and crop a region of $200 \times 200$ pixels centered at the detected window.

\subsubsection{Fingerprint Images}
Given the diverse nature of images captured from different sensors, here the preprocessing is defined according to the sensor type.
\begin{enumerate}[(a)]
\item \emph{Biometrika}: we cropped the central region of size in columns and rows corresponding to 70\% of the original image dimensions. 
\item \emph{Italdata} and \emph{CrossMatch}: we cropped the central region of size in columns and rows respectively corresponding to 60\% and 90\% of the original image columns and rows.
\item \emph{Swipe}: As the images acquired by this sensor contain a variable number of blank rows at the bottom, the average number of non-blank rows $M$ was first calculated from the training images.
Then, in order to obtain images of a common size with non-blank rows, we removed their blank rows at the bottom and rescaled them to $M$ rows. Finally, we cropped the central region corresponding to 90\% of original image columns and $M$ rows.
\end{enumerate}

The rationale for these operations is based on the observation that fingerprint images in LivDet2013 tend to have a large portion of background content and therefore we try to discard such information that could otherwise mislead the representation learning process.
The percentage of cropped columns and rows differs among sensors because they capture images of different sizes with different amounts of background.

For architecture optimization (AO), the decision to use image color information was made according to 10-fold validation (see Section~\ref{sec:ao}), while for filter optimization (FO), color information was considered whenever available for a better approximation with the standard cf10-11 architecture. Finally, images were resized to $32\times32$ or $128\times128$ to be taken as input for the cf10-11 and spoofnet architectures, respectively. 

\subsection{Evaluation Protocol}
\label{sec:evalprot}

For each benchmark, we learn deep representations from their training images according to the methodology described in Section~\ref{sec:ao} for architecture optimization (AO) and in Section~\ref{sec:fo} for filter optimization (FO).
We follow the standard evaluation protocol of all benchmarks and evaluate the methods in terms of detection accuracy (ACC) and half total error rate (HTER), as these are the metrics used to assess progress in the set of benchmarks considered herein. Precisely, for a given benchmark and convolutional network already trained, results are obtained by:

\begin{enumerate}
\item Retrieving prediction scores from the testing samples;
\item Calculating a threshold $\tau$ above which samples are predicted as attacks;
\item Computing ACC and/or HTER using $\tau$ and test predictions.
\end{enumerate}

The way that $\tau$ is calculated differs depending on whether the benchmark has a development set or not (Table~\ref{tab:databases}). Both face benchmarks have such a set and, in this case, we simply obtain $\tau$ from the predictions of the samples in this set. 
Iris and fingerprint benchmarks have no such a set, therefore $\tau$ is calculated depending on whether the convolutional network was learned with AO or FO.

In case of AO, we calculate $\tau$ by joining the predictions obtained from 10-fold validation (see Section~\ref{sec:ao}) in a single set of positive and negative scores, and $\tau$ is computed as the point that lead to an equal error rate (EER) on the score distribution under consideration. 
In case of FO, scores are probabilities and we assume $\tau=0.5$. ACC and HTER are then trivially computed with $\tau$ on the testing set.

It is worth noting that the Warsaw iris benchmark provides a supplementary testing set that here we merge with the original testing set in order to replicate the protocol of~\cite{LivDet:Iris:2013}.
Moreover, given face benchmarks are video-based and that in our methodology we treat them as images (Section~\ref{sec:preproc}),
we perform a score-level fusion of the samples from the same video according to the max rule~\cite{Ross:HM:2006}. This fusion is done before calculating $\tau$.

\subsection{Implementation}
\label{sec:implementationdetais}

Our implementation for architecture optimization (AO) is based on Hyperopt-convnet~\cite{Bergstra:2013b} which in turn is based on Theano~\cite{Bergstra:SCIPY:2010}.
LibSVM~\cite{Chang:2011} is used for learning the linear classifiers via Scikit-learn.\footnote{http://scikit-learn.org}
The code for feature extraction runs on GPUs due to Theano and the remaining part is multithreaded and runs on CPUs.
We extended Hyperopt-convnet in order to consider the operations and hyperparameters as described in Appendix~\ref{sec:convnet_ops} and Section~\ref{sec:ao} and we will make the source code freely available in~\cite{Chiachia:2014b}.
Running times are reported with this software stack and are computed in an Intel i7 @3.5GHz with a Tesla K40 that, on average, takes less than one day to optimize an architecture --- i.e., to probe 2,000 candidate architectures --- for a given benchmark.

As for filter optimization (FO), Cuda-convnet~\cite{Krizhevsky:cuda-convnet:2012} is used. 
This library has an extremely efficient implementation to train convolutional networks via back-propagation on NVIDIA GPUs. 
Moreover, it provides us with the cf10-11 convolutional architecture taken in this work as reference for FO.

\begin{table*}[htb]
\begin{center}
\caption{
Overall results considering relevant information of the best found architectures, detection accuracy (ACC) and HTER values according to the evaluation protocol, and state-of-the-art (SOTA) performance.
}
\label{tab:resultsOARF}
\iffinal
\else
\scriptsize
\fi
\begin{tabular}{llr@{}c@{}l@{ }c@{ }c@{ }r@{ }c@{}l@{ }ccc@{ }ccc@{ }c@{ }c@{}c}
\hline
\multirow{3}{*}{modality}
& \multirow{3}{*}{benchmark} 
                & \multicolumn{9}{c}{architecture optimization (AO)}                     && \multicolumn{2}{c}{our results} && \multicolumn{3}{c}{SOTA results} & \\
                \cline{3-11} 
                \cline{13-14} \cline{16-18} 
&               &  \multicolumn{3}{c}{time}
                           & size 
                                 & layers
                                     & \multicolumn{3}{c}{features} 
                                                                         & objective 
                                                                                  && ACC & HTER && ACC & HTER & \multirow{2}{*}{Ref.}\\
&               & \multicolumn{3}{c}{(secs.)}
                          & (pixels) 
                                 &$(\#)$
                                     & \multicolumn{3}{c}{$(\#)$} 
                                                                          &  $(\%)$    &&$(\%)$ &$(\%)$ &&$(\%)$  & $(\%)$ & \\
\hline\hline
\multirow{2}{*}{iris}
& Warsaw            &  52&+&35 & 640 & 2 & $10\times 15\times  64$ &&  (9600) & 98.21 && \textbf{99.84}
                                                                                            &  0.16 && 97.50  & ---    & \cite{Czajka:MMAR:2013} \\ 
& Biosec        &  80&+&34 & 640 & 3 & $ 2\times  5\times 256$ &&  (2560) & 97.56 &&  98.93 &  1.17 && \textbf{100.00}
                                                                                                              & ---    & \cite{Galbally:ICB:2012} \\ 
& MobBIOfake    &  18&+&37 & 250 & 2 & $ 5\times  7\times 256$ &&  (8960) & 98.94 &&  98.63 &  1.38 && \textbf{99.75}
                                                                                                                       & ---    & \cite{Sequeira:IJCB:2014} \\ 
\hline                                      
\multirow{2}{*}{face}
& Replay-Attack &  69&+&15 & 256 & 2 & $ 3\times  3\times 256$ &&  (2304) & 94.65 &&  98.75 &  \textbf{0.75} && ---    &   5.11 & \cite{Komulainen:ICB:2013} \\ 
& 3DMAD         &  55&+&15 & 128 & 2 & $ 5\times  5\times  64$ &&  (1600) & 98.68 && 100.00 &  \textbf{0.00}
                                                                                                    && ---    &   0.95 & \cite{Erdogmus:BIOSIG:2013}\\ 
\hline
\multirow{2}{*}{fingerprint}
& Biometrika    &  66&+&25 & 256 & 2 & $ 2\times  2\times 256$ &&  (1024) & 90.11 &&  96.50 &  3.50 &&  \textbf{98.30}
                                                                                                                       & ---    & \cite{Ghiani:ICB:2013} \\ 
& Crossmatch    & 112&+&12 & 675 & 3 & $ 2\times  3\times 256$ &&  (1536) & 91.70 && \textbf{92.09} 
                                                                                            &  8.44 &&  68.80 & ---    & \cite{Ghiani:ICB:2013} \\ 
& Italdata      &  46&+&27 & 432 & 3 & $16\times 13\times 128$ && (26624) & 86.89 &&  97.45 &  2.55 &&  \textbf{99.40}
                                                                                                                       & ---    & \cite{Ghiani:ICB:2013} \\ 
& Swipe         &  97&+&51 & 962 & 2 & $53\times  3\times  32$ &&  (5088) & 90.32 &&  88.94 & 11.47 &&  \textbf{96.47}
                                                                                                                       & ---    & \cite{Ghiani:ICB:2013} \\ 
\hline
\end{tabular}
\end{center}
\end{table*}

\section{Experiments and Results}
\label{sec:experiments}

In this section, we evaluate the effectiveness of the proposed methods for spoofing detection. We show experiments for the architecture optimization and filter learning approaches along with their combination  
for detecting iris, face, and fingerprint spoofing on the nine benchmarks described in Section~\ref{sec:databases}. We also present results for the \emph{spoofnet}, which incorporates some domain-knowledge on the problem. We compare all of the results with the state-of-the-art counterparts. Finally, we discuss the pros and cons of using such approaches and their combination along with efforts to understand the type of features learned and some effeciency questions when testing the proposed methods.

\subsection{Architecture Optimization (AO)}

Table~\ref{tab:resultsOARF} presents AO results in detail as well as previous state-of-the-art (SOTA) performance for the considered benchmarks. With this approach, we can outperform four SOTA methods in all three biometric modalities. Given that AO assumes little knowledge about the problem domain, this is remarkable. Moreover, performance is on par in other four benchmarks, with the only exception of Swipe. Still in Table~\ref{tab:resultsOARF}, we can see information about the best architecture such as time taken to evaluate it (feature extraction + 10-fold validation), input size, depth, and dimensionality of the output representation in terms of \emph{columns} $\times$ \emph{rows} $\times$ \emph{feature maps}.

Regarding the number of layers in the best architectures, we can observe that six out of nine networks use two layers, and three use three layers. We speculate that the number of layers obtained is a function of the problem complexity.
In fact, even though there are many other hyperparameters involved, the number of layers play an important role in this issue, since it directly influences the level of non-linearity and abstraction of the output with respect to the input.

With respect to the input size, we can see in comparison with Table~\ref{tab:databases:experiments}, that the best performing architectures often use the original image size. This was the case for all iris benchmarks and for three (out of four) fingerprint benchmarks. For face benchmarks, a larger input was preferred for Replay-Attack, while a smaller input was preferred for 3DMAD. We hypothesize that this is also related to the problem difficulty, given that Replay-Attack seems to be more difficult, and that larger inputs tend to lead to larger networks.

We still notice that the dimensionality of the obtained representations are, in general, smaller than 10K features, except for Italdata.
Moreover, for the face and iris benchmarks, it is possible to roughly observe a relationship between the optimization objective calculated in the training set and the detection accuracy measure on the testing set (Section~\ref{sec:evalprot}), which indicates the appropriateness of the objective for these tasks. However, for the fingerprint benchmarks, this relationship does not exist, and we accredit this to either a deficiency of the optimization objective in modelling these problems or to the existence of artifacts in the training set misguiding the optimization.

\subsection{Filter Optimization (FO)}

\begin{table}
\begin{center}
\caption{
Results for filter optimization (FO) in \emph{cf10-11} and \emph{spoofnet} (Fig.~\ref{fig:cudaconvnet}). Even though both networks present similar behavior, \emph{spoofnet} is able to push performance even further in problems which \emph{cf10-11} was already good for.
Architecture optimization (AO) results (with random filters) are shown in the first column to facilitate comparisons.
}
\label{tab:FO}
\begin{tabular}{c@{ }c@{ }l@{ }r@{ }r@{ }r@{ }r@{ }r@{ }r@{ }r@{ }r@{ }c}
\hline
         & \hspace{1em} && \multicolumn{6}{c}{filter}\\ \cline{4-9}
modality & \hspace{1em} &&& random && \multicolumn{3}{c}{optimized} \\ \cline{5-5} \cline{7-9}
(metric) && benchmark \hspace{1em} && \multicolumn{1}{c}{AO} && \textit{cf10-11} && \textit{spoofnet} && SOTA & \\
\hline
iris && Warsaw        && \textbf{99.84}  && 67.20 &&          66.42  &&          97.50 \\
(ACC)&& Biosec        &&         98.93   && 59.08 &&          47.67  && \textbf{100.00} \\
     && MobBIOfake    &&         98.63   && 99.13 && \textbf{100.00} &&          99.75 \\
\hline
face && Replay-Attack &&\textbf{0.75} && 55.13 && 55.38 && 5.11\\
(HTER) && 3DMAD         &&\textbf{0.00} && 40.00 && 24.00 && 0.95\\ 
\hline
fingerprint &&Biometrika    && 96.50 && 98.50&& \textbf{99.85} && 98.30\\
(ACC)     &&Crossmatch    && 92.09 && 97.33&& \textbf{98.23} && 68.80\\
     &&Italdata      && 97.45 && 97.35&& \textbf{99.95} && 99.40\\
     &&Swipe         && 88.94 && 98.70&& \textbf{99.08} && 96.47\\
\hline
\end{tabular}
\end{center}
\end{table}

Table~\ref{tab:FO} shows the results for FO, where we repeat architecture optimization (AO) results (with random filters) in the first column to facilitate comparisons. Overall, we can see that both networks, \emph{cf10-11} and \emph{spoofnet} have similar behavior across the biometric modalities.

Surprisingly, \emph{cf10-11} obtains excellent performance in all four fingerprint benchmarks as well as in the MobBIOFake, exceeding SOTA in three cases, in spite of the fact that it was used without any modification. However, in both face problems and in two iris problems, \emph{cf10-11} performed poorly. Such difference in performance was not possible to anticipate by observing training errors, which steadily decreased in all cases until training was stopped. Therefore, we believe that in these cases FO was misguided by the lack of training data or structure in the training samples irrelevant to the problem.

To reinforce this claim, we performed experiments with filter optimization (FO) in \emph{spoofnet} by varying the training set size with 20\%, 40\%, and 50\% of fingerprint benchmarks. As expected, in all cases, the less training examples, the worse is the generalization of the \emph{spoofnet} (lower classification accuracies). 
Considering the training phase, for instance, when using 50\% of training set or less, the accuracy achieved by the learned representation is far worse than the one achieved when using 100\% of training data. 
This fact reinforces the conclusion presented herein regarding the small sample size problem. 
Maybe a fine-tuning of some parameters, such as the number of training epochs and the learning rates, can diminish the impact of the small sample size problem stated here, however, this is an open research topic by itself.

For \emph{spoofnet}, the outcome is similar. As we expected, the proposed architecture was able to push performance even further in problems which \emph{cf10-11} was already good for, outperforming SOTA in five out of nine benchmarks. This is possibly because we made the \emph{spoofnet} architecture simpler, with less parameters, and taking input images with a size better suited to the problem.

As compared to the results in AO, we can observe a good balance between the approaches. In AO, the resulting convolutional networks are remarkable in the face benchmarks. In FO, networks are remarkable in fingerprint problems. While in AO all optimized architectures have good performance in iris problems, FO excelled in one of these problems, MobBIOFake, with a classification accuracy of 100\%. In general, AO seems to result in convolutional networks that are more stable across the benchmarks, while FO shines in problems in which learning effectively occurs. Considering both AO and FO, we can see in Table~\ref{tab:FO} that we outperformed SOTA methods in eight out of nine benchmarks. The only benchmark were SOTA performance was not achieved is Biosec, but even in this case the result obtained with AO is competitive. 

\begin{figure}
\centering
\subfloat[Filter weights of the first convolutional layer that were learned using the backpropagation algorithm.]{\includegraphics[width=0.46\textwidth]{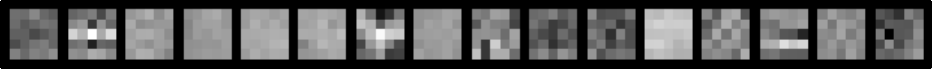}\label{fig:filter_learned_conv1}}\\
\subfloat[Mean for real images (test set).]{\includegraphics[width=0.20\textwidth]{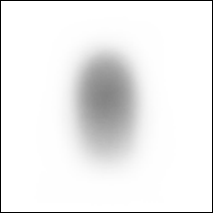}\label{fig:mean_pos_class}}\hspace{1mm}
\subfloat[Mean for fake images (test set).]{\includegraphics[width=0.20\textwidth]{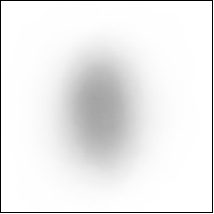}\label{fig:mean_neg_class}}\\
\subfloat[Activation maps for real images.]{\includegraphics[width=0.23\textwidth]{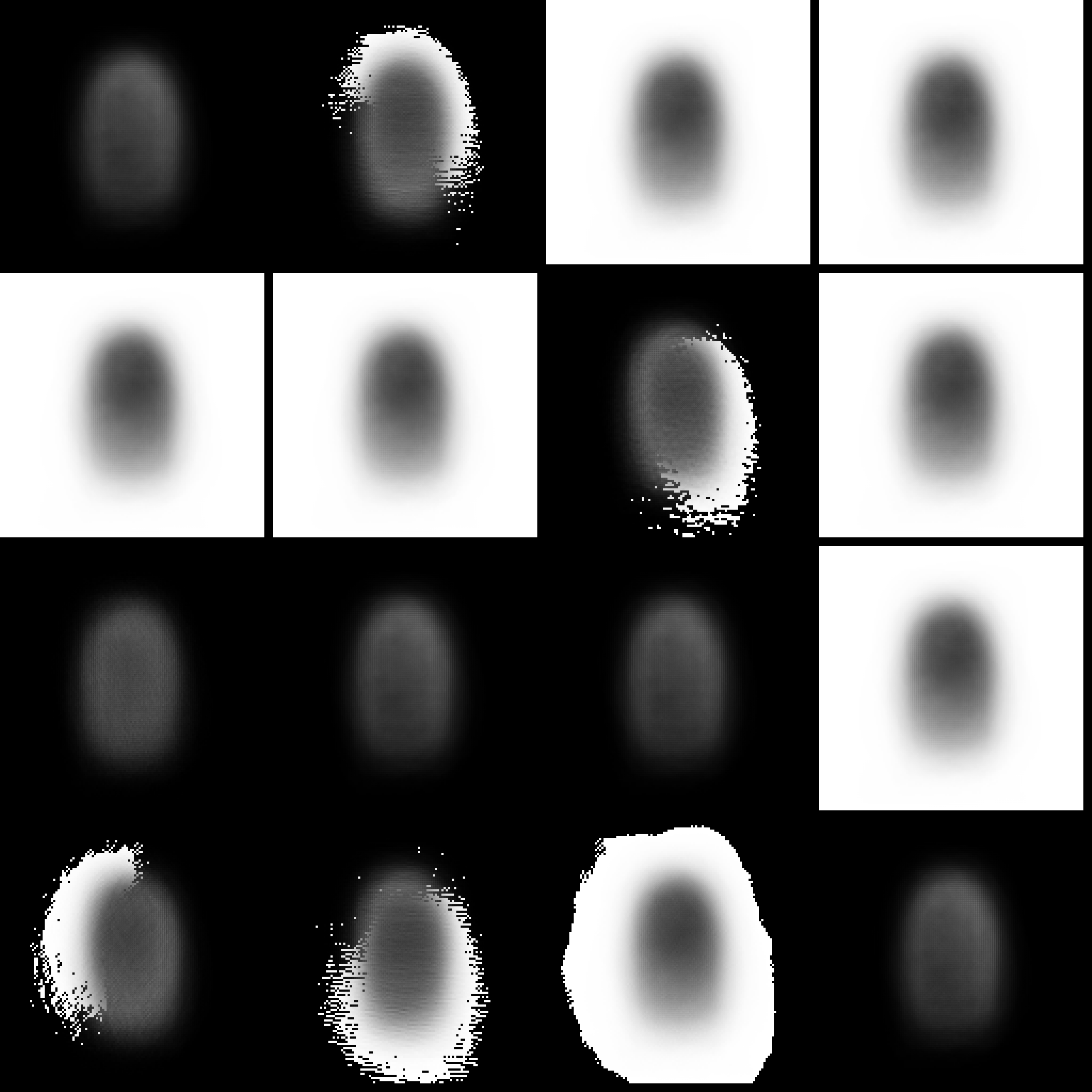}\label{fig:act_map_class_1}}\hspace{1mm}
\subfloat[Activation maps for fake images.]{\includegraphics[width=0.23\textwidth]{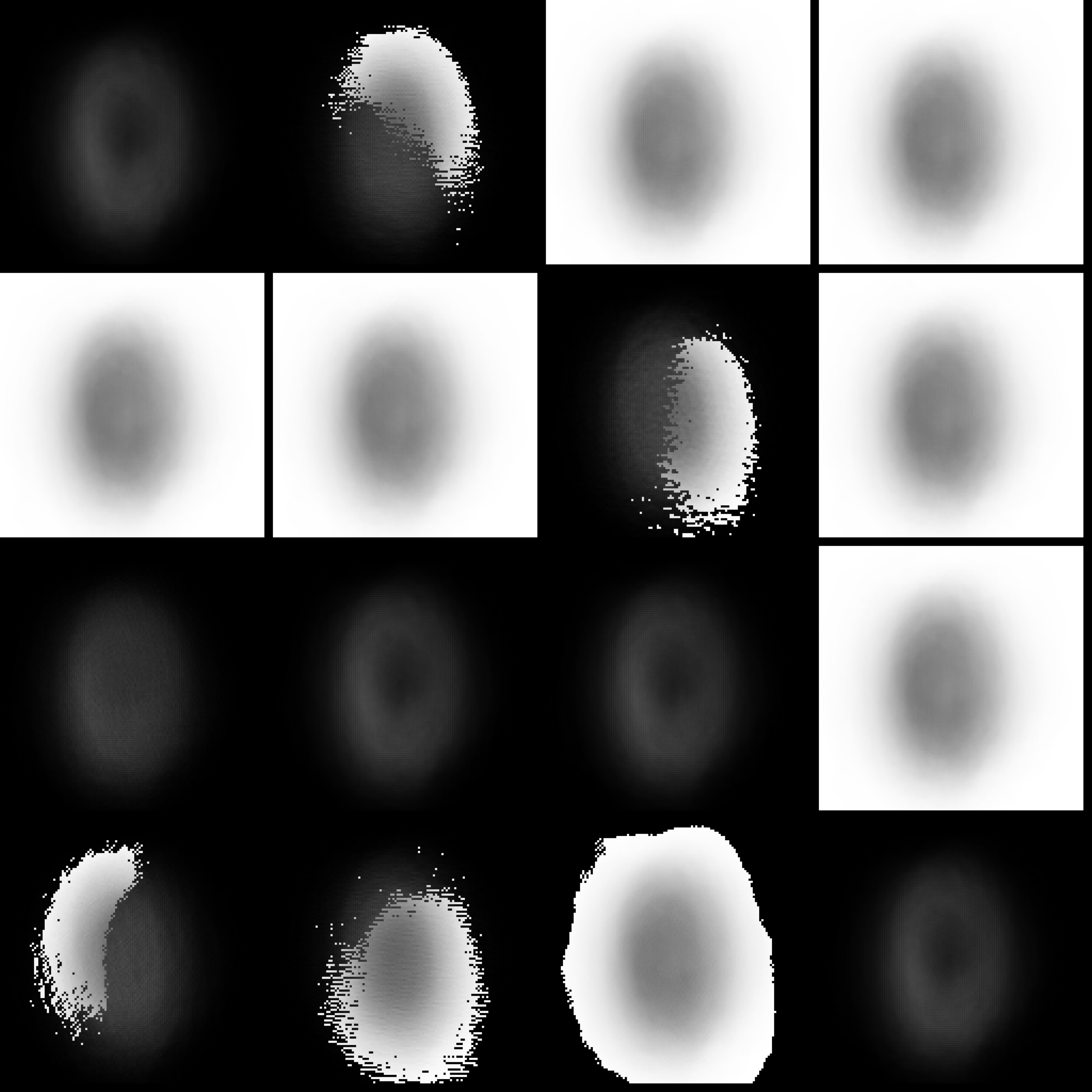}\label{fig:act_map_class_0}}\\

\caption{Activation maps of the filters that compose the first convolutional layer when forwarding real and fake images through the network.}
\end{figure}

Understanding how a set of deep learned features capture properties and nuances of a problem is still an open question in the vision community. However, in an attempt to understand the behavior of the operations applied onto images after they are forwarded through the first convolutional layer, we generate Fig.~\ref{fig:filter_learned_conv1} that illustrates the filters learned via backpropagation algorithm and Figs.~\ref{fig:mean_pos_class}~and~\ref{fig:mean_neg_class} showing the mean of real and fake images that compose the test set, respectively. To obtain output values from the first convolutional layer and get a sense of them, we also instrumented the \textit{spoofnet} convolutional network to forward the real and fake images from the test set through network. Figs~\ref{fig:act_map_class_1}~and~\ref{fig:act_map_class_0} show such images for the real and fake classes, respectively.

We can see in Fig.~\ref{fig:filter_learned_conv1} that the filters learned patterns resemble textural patterns instead of edge patterns as usually occurs in several computer vision problems~\cite{Krizhevsky:2012,Ouyang:2014}. This is particularly interesting and in line with several anti-spoofing methods in the the literature which also report good results when exploring texture information~\cite{Ghiani:ICB:2013, Maatta:IJCB:2011}.

In addition, Fig.~\ref{fig:mean_pos_class}~and~\ref{fig:mean_neg_class} show there are differences between real and fake images from test, although apparently small in such a way that a direct analysis of the images would not be enough for decision making. However, when we analyze the mean activation maps for each class, we can see more interesting patterns. In Figs.~\ref{fig:act_map_class_1}~and~\ref{fig:act_map_class_0}, we have sixteen pictures with $128 \times 128$ pixel resolution. These images correspond to the sixteen filters that composing the first layer of the \textit{spoofnet}. Each position $(x,y)$ in these $128 \times 128$ images corresponds to a $5 \times 5$ area (receptive field units) in the input images. Null values in a given unit means that the receptive field of the unit was not able to respond to the input stimuli. In contrast, non-null values mean that the receptive field of the unit had a responsiveness to the input stimuli.

We can see that six filters have a high responsiveness to the background information of the input images (filters predominantly white) whilst ten filters did not respond to background information (filters predominantly black). From left to right, top to bottom, we can see  also that the images corresponding to the filters 2, 7, 13, 14 and 15 have high responsiveness to information surrounding the central region of the sensor (usually where fingerprints are present) and rich in texture datails. Although these regions of high and low responsiveness are similar for both classes we can notice some differences. A significant difference in this first convolutional layer to images for the different classes is that the response of the filters regarding to fake images (Fig~\ref{fig:act_map_class_0}) generates a blurring pattern, unlike the responses of the filters regarding to real images (Fig~\ref{fig:act_map_class_1}) which generate a sharper pattern. We believe that the same way as the first layer of a convolutional network has the ability to respond to simple and relevant patterns (edge information) to a problem of recognition objects in general, in computer vision, the first layer in the \textit{spoofnet} also was able to react to a simple pattern recurrent in spoof problems, the blurring effect, an artifact previously explored in the literature~\cite{Galbally:TIP:2014}. Finally, we are exploring visualisation only of the first layer; subsequent layers of the network can find new patterns in these regions activated by the first layer further emphasizing class differences. 

\subsection{Interplay between AO and FO}

\begin{table}
\begin{center}
\caption{Results for architecture and filter optimization (AO+FO) along with \emph{cf10-11} and \emph{spoofnet} networks considering random weights.
AO+FO show compelling results for fingerprints and one iris benchmark (MobBIOFake). We can also see that \emph{spoofnet} can benefit from random filters in situations it was not good for when using filter learning (e.g., Replay-Attack).}
\label{tab:interplay}
\begin{tabular}{c@{ }c@{ }l@{ }r@{ }c@{ }r@{ }r@{ }r@{ }r@{ }r@{ }r@{ }c}
\hline
         & \hspace{1em} && \multicolumn{6}{c}{filter}\\ \cline{4-9}
modality & \hspace{1em} &&& optimized && \multicolumn{3}{c}{random} \\ \cline{5-5} \cline{7-9}
(metric) && benchmark \hspace{1em} && AO && \textit{cf10-11} && \textit{spoofnet} && SOTA & \\
\hline
iris && Warsaw        &&                59.55  && 87.06 &&          96.44  && \textbf{ 97.50} \\
(ACC)&& Biosec        &&                57.50  && 97.33 &&          97.42  && \textbf{100.00} \\
     && MobBIOfake    &&                99.38  && 77.00 &&          72.00  &&  \textbf{ 99.75} \\
\hline
face && Replay-Attack &&                55.88  &&  5.62 &&  \textbf{ 3.50} &&           5.11 \\
(HTER) && 3DMAD       &&                40.00  &&  8.00 &&           4.00  &&  \textbf{ 0.95} \\ 
\hline
fingerprint &&Biometrika     && \textbf{99.30}  && 77.45 &&          94.70  &&           98.30\\
(ACC)     &&Crossmatch       && \textbf{98.04}  && 83.11 &&          87.82  &&           68.80\\
     &&Italdata              && \textbf{99.45}  && 76.45 &&          91.05  &&           99.40\\
     &&Swipe                 && \textbf{99.08}  && 87.60 &&          96.75  &&           96.47\\
\hline
\end{tabular}
\end{center}
\end{table}

In the previous experiments, architecture optimization (AO) was evaluated using random filters and filter optimization (FO) was carried out in the predefined architectures \emph{cf10-11} and \emph{spoofnet}. A natural question that emerges in this context is how these methods would perform if we (i) combine AO and FO and if we (ii) consider random filters in \emph{cf10-11} and \emph{spoofnet}.

Results from these combinations are available in Table~\ref{tab:interplay} and show a clear pattern. When combined with AO, FO again exceeds previous SOTA in all fingerprint benchmarks and performs remarkably good in MobBIOFake. However, the same difficulty found by FO in previous experiments for both face and two iris benchmarks is also observed here.
Even though \emph{spoofnet} performs slightly better than AO in the cases where SOTA is exceeded (Table~\ref{tab:FO}), it is important to remark that our AO approach may result in architectures with a much larger number of filter weights to be optimized, and this may have benefited~\emph{spoofnet}.

It is also interesting to observe in Table~\ref{tab:interplay} the results obtained with the use of random filters in \emph{cf10-11} and \emph{spoofnet}. The overall balance in performance of both networks across the benchmarks is improved, similar to what we have observed with the use of random filters in Table~\ref{tab:resultsOARF}. An striking observation is that~\emph{spoofnet} with random filters exceed previous SOTA in Replay-Attack, and this supports the idea that the poor performance of~\emph{spoofnet} in Replay-Attack observed in the FO experiments (Table~\ref{tab:FO}) was not a matter of architecture.

\subsection{Runtime}

We estimate time requirements for anti-spoofing systems built with convolutional networks based on measurements obtained in architecture optimization (AO).
We can see in~Table~\ref{tab:resultsOARF} that the most computationally intensive deep representation is the one found for the Swipe benchmark, and demands 148 (97+51) seconds to process 2,200 images. Such a running time is only possible due to the GPU+CPU implementation used (Section~\ref{sec:implementationdetais}), which is critical for this type of learning task. In a hypothetical operational scenario, we could ignore the time required for classifier training (51 seconds, in this case). Therefore, we can estimate that, on average, a single image captured by a Swipe sensor would require approximately 45 milliseconds --- plus a little overhead --- to be fully processed in this hypothetical system. Moreover, the existence of much larger convolutional networks running in realtime in budgeted mobile devices~\cite{Wardern:2014} also supports the idea that the approach is readily applicable in a number of possible scenarios.

\begin{figure*}[!t]
\begin{center}
\includegraphics[width=0.85\linewidth]{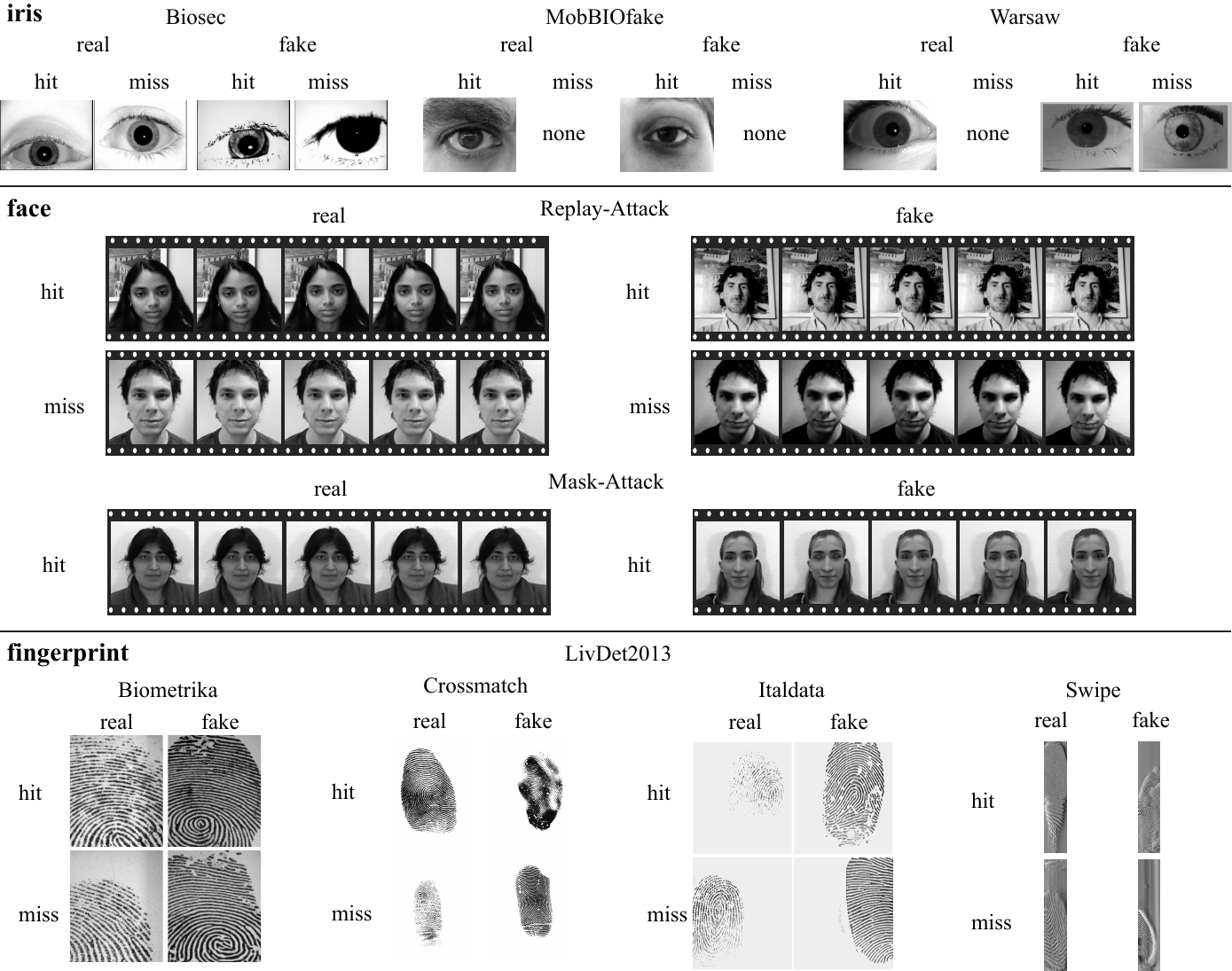}
\caption{Examples of hit and missed testing samples lying closest to the real-fake decision boundary of each benchmark. A magnified visual inspection on these images may suggest some properties of the problem to which the learned representations are sensitive.}
\label{fig:datasets}\label{page:fig:datasets}
\end{center}
\end{figure*}

\subsection{Visual Assessment}

In Fig.~\ref{fig:datasets}, we show examples of hit and missed testing samples lying closest to the real-fake decision boundary of the best performing system in each benchmark. 
A magnified visual inspection on these images may give us some hint about properties of the problem to which the learned representations are sensitive.

While it is difficulty to infer anything concrete, it is interesting to see that the real missed sample in Biosec is quite bright, and that skin texture is almost absent in this case. 
Still, we may argue that a noticeable difference exists in Warsaw between the resolution used to print the images that led to the fake hit and the fake miss.

Regarding the face benchmarks, the only noticeable observation from Replay-Attack is that the same person is missed both when providing to the system a real and a fake biometric reading. This may indicate that some individuals are more likely to successfully attack a face recognition systems than others. In 3DMAD, it is easy to see the difference between the real and fake hits. Notice that there was no misses in this benchmark.

A similar visual inspection is much harder in the fingerprint benchmarks, even though the learned deep representations could effectively characterize these problems. The only observation possible to be made here is related to the fake hit on CrossMatch, which is clearly abnormal.
The images captured with the Swipe sensor are naturally narrow and distorted due to the process of acquisition, and this distortion prevents any such observation.
\section{Conclusions and Future Work}
\label{sec:conclusions}

In this work, we investigated two deep representation research approaches for detecting spoofing in different biometric modalities. On one hand, we approached the problem by learning representations directly from the data through architecture optimization with a final decision-making step atop the representations. On the other, we sought to learn filter weights for a given architecture using the well-known back-propagation algorithm. As the two approaches might seem naturally connected, we also examined their interplay when taken together. In addition, we incorporated our experience with architecture optimization as well as with training filter weight for a given architecture into a more interesting and adapted network, \emph{spoofnet}.  

Experiments showed that these approaches achieved outstanding classification results for all problems and modalities outperforming the state-of-the-art results in eight out of nine benchmarks.
Interestingly, the only case for which our approaches did not achieve SOTA results is for the Biosec benchmark. However, in this case, it is possible to achieve a 98.93\% against 100.0\% accuracy of the literature.
These results support our hypothesis that the conception of data-driven systems using deep representations able to extract semantic and vision meaningful features directly from the data is a promising venue. Another indication of this comes from the initial study we did for understanding the type of filters generated by the learning process. Considering the fingerprint case, learning directly from data, it was possible to come up with discriminative filters that explore the blurring artifacts due to recapture. This is particularly interesting as it is in line with previous studies using custom-tailored solutions~\cite{Galbally:TIP:2014}.

It is important to emphasise the interplay between the architecture and filter optimization approaches for the spoofing problem.
It is well-known in the deep learning literature that when thousands of samples are available for learning, the filter learning approach is a promising path. Indeed, we could corroborate this through fingerprint benchmarks that considers a few thousand samples for training. However, it was not the case for faces and two iris benchmarks which suffer from the small sample size problem (SSS) and subject variability hindering the filter learning process. In these cases, the architecture optimization approach was able to learn representative and discriminative features providing comparable spoofing effectiveness to the SOTA results in almost all benchmarks, and specially outperforming them in three out of four SOTA results when the filter learning approach failed. It is worth mentioning that sometimes it is still possible to learn meaningful features from the data even with a small sample size for training. We believe this happens in more well-posed datasets with less variability between training/testing data as it is the case of MobioBIOfake benchmark in which the AO approach achieved 99.38\% just 0.37\% behind the SOTA result. 

As the data tell it all, the decision to which path to follow can also come from the data. Using the evaluation/validation set during training, the researcher/developer can opt for optimizing architectures, learn filters or both. If training time is an issue and a solution must be presented overnight, it might be interesting to consider an already learned network that incorporates some additional knowledge in its design. In this sense, \emph{spoofnet} could be a good choice. In all cases, if the developer can incorporate more training examples, the approaches might benefit from such augmented training data. 


The proposed approaches can also be adapted to other biometric modalities not directly dealt with herein. The most important difference would be in the input type of data since all discussed solutions directly learn their representations from the data.

For the case of iris spoofing detection, here we dealt only with iris spoofing printed attacks and some experimental datasets using cosmetic contact lenses have recently become available allowing researchers to study this specific type of spoofing~\cite{Bowyer:Computer:2014,Yadav:TIFS:2014}. 
For future work, we intend to evaluate such datasets using the proposed approaches here and also consider other biometric modalities such as palm, vein, and gait.

Finally, it is important to take all the results discussed herein with a grain of salt. We are not presenting the final word in spoofing detection. In fact, there are important additional research that could finally take this research another step forward. We envision the application of deep learning representations on top of pre-processed image feature maps (e.g., LBP-like feature maps, acquisition-based maps exploring noise signatures, visual rhythm representations, etc.). With an $n$-layer feature representation, we might be able to explore features otherwise not possible using the raw data. In addition, exploring temporal coherence and fusion would be also important for video-based attacks.

\appendices

\section{Convolutional Network Operations}
\label{sec:convnet_ops}

Our networks use classic convolutional operations that can be viewed as linear and non-linear image processing operations. When stacked, these operations essentially extract higher level representations, named \emph{multiband images}, whose pixel attributes are concatenated into high-dimensional feature vectors for later pattern recognition.\footnote{This appendix describes convolutional networks from an image processing perspective, therefore the use of terms like image \emph{domain}, image \emph{band}, \emph{etc.}}

Assuming $\hat{I}=(D_I,\vec{I})$ as a multiband image, where $D_I \subset Z^2$ is the image domain and $\vec{I}(p)=\{I_1(p),I_2(p),\ldots, I_m(p)\}$ is the attribute vector of a $m$-band pixel $p=(x_p,y_p)\in D_I$, the aforementioned operations can be described as follows.

\subsubsection{Filter Bank Convolution}

Let ${\cal A}(p)$ be a squared region centered at $p$ of size $L_{\cal A} \times L_{\cal A}$, such that ${\cal A} \subset D_I$ and $q\in {\cal A}(p)$ iff $\max(|x_q-x_p|,|y_q-y_p|)\leq (L_{\cal A}-1)/2$.
Additionally, let $\Phi=({\cal A},W)$ be a filter with weights $W(q)$ associated with pixels $q\in {\cal A}(p)$.
In the case of multiband filters, filter weights can be represented as vectors $\vec{W}_i(q)=\{w_{i,1}(q),w_{i,2}(q),\ldots,w_{i,m}(q)\}$ for each filter $i$ of the bank, and a multiband filter bank $\Phi = \{\Phi_1, \Phi_2, \ldots, \Phi_n\}$ is a set of filters $\Phi_i=({\cal A},\vec{W}_i)$, $i=\{1,2,\ldots,n\}$.

The convolution between an input image $\hat{I}$ and a filter $\Phi_i$ produces a band $i$ of the filtered image $\hat{J}=(D_J,\vec{J})$, where $D_J \subset D_I$ and $\vec{J}=(J_1,J_2,\ldots,J_n)$, such that for each $p\in D_J$,
\begin{equation}
J_i(p) = \sum_{\forall q\in {\cal A}(p)} \vec{I}(q)\cdot \vec{W}_i(q).
\end{equation}

\subsubsection{Rectified Linear Activation}

Filter activation in this work is performed by rectified linear units (RELUs) of the type present in many state-of-the-art convolutional architectures~\cite{Krizhevsky:2012,Pinto:2011b} and is defined as
\begin{equation}
J_i(p) = \max(J_i(p),0).
\end{equation}

\subsubsection{Spatial Pooling}

Spatial pooling is an operation of paramount importance in the literature of convolutional networks~\cite{LeCun:1998} that aims at bringing translational invariance to the features by aggregating activations from the same filter in a given region. 

Let ${\cal B}(p)$ be a pooling region of size $L_{\cal B} \times L_{\cal B}$ centered at pixel $p$ and $D_K = D_J/s$ be a regular subsampling of every $s$ pixels $p \in D_J$. We call $s$ the \emph{stride} of the pooling operation. Given that $D_J \subset Z^2$, if $s=2$, $|D_K| = |D_J|/4$, for example.
The pooling operation resulting in the image $\hat{K}=(D_K,\vec{K})$  is defined as 
\begin{equation}
K_i(p) = \sqrt[\alpha]{\sum_{\forall q\in {\cal B}(p)} J_i(q)^{\alpha}}, 
\end{equation}
where $p\in D_K$ are pixels in the new image, $i=\{1,2,\ldots,n\}$ are the image bands, and $\alpha$ is a hyperparameter that controls the sensitivity of the operation. In other words, our pooling operation is the $L_{\alpha}$-norm of values in ${\cal B}(p)$. 
The stride $s$ and the size of the pooling neighborhood defined by $L_{\cal B}$ are other hyperparameters of the operation.

\subsubsection{Divisive Normalization}

The last operation considered in this work is divisive normalization, a mechanism widely used in top-performing convolutional networks~\cite{Krizhevsky:2012,Pinto:2011b} that is based on gain control mechanisms found in cortical neurons~\cite{Geisler:1992}.

This operation is also defined within a squared region ${\cal C}(p)$ of size $L_{\cal C} \times L_{\cal C}$ centered at pixel $p$ such that 
\begin{equation}
O_i(p) = \frac{K_i(p)}{\sqrt{\sum_{j=1}^{n}\sum_{\forall q\in {\cal C}(p)} K_j(q)^2}}
\end{equation}
for each pixel $p\in D_O \subset D_K$ of the resulting image $\hat{O}=(D_O,\vec{O})$. Divisive normalization promotes competition among pooled filter bands such that high responses will prevail even more over low ones, further strengthening the robustness of the output representation $\vec{O}$.

\section*{Acknowledgment}

We thank UFOP, Brazilian National Research Counsil -- CNPq (Grants \#303673/2010-9,  \#304352/2012-8, \#307113/2012-4, \#477662/2013-7, \#487529/2013-8, \#479070/2013-0, and \#477457/2013-4), the CAPES DeepEyes project, S\~ao Paulo Research Foundation -- FAPESP, (Grants \#2010/05647-4, \#2011/22749-8, \#2013/04172-0, and \#2013/11359-0), and Minas Gerais Research Foundation -- FAPEMIG (Grant APQ-01806-13). 
D. Menotti thanks FAPESP for a grant to acquiring two NVIDIA GeForce GTX Titan Black with 6GB each.
We also thank NVIDIA for donating five GPUs used in the experiments, a Tesla K40 with 12GB to A. X. Falc{\~a}o, two GeForce GTX 680 with 2GB each to G. Chiachia, and two GeForce GTX Titan Black with 6GB each to D. Menotti.

\ifCLASSOPTIONcaptionsoff
  \newpage
\fi


\bibliographystyle{IEEEtran}
\bibliography{spoofing,iris,faces,fingerprint} 

\begin{IEEEbiography}[{\includegraphics[width=1in,height=1.25in,clip,keepaspectratio]{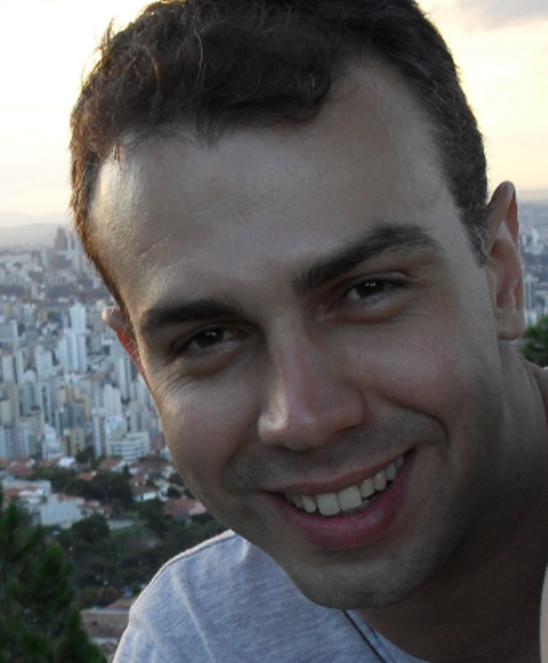}}]{David Menotti} received his Computer Engineering and Informatics Applied Master degrees from the Pontif\'icia Universidade Cat\'olica do Paran\'a (PUCPR), Curitiba, Brazil, in 2001 and 2003, respectively.
In 2008, he received his co-tutelage PhD degree in Computer Science from the Federal University of Minas Gerais (UFMG), Belo Horizonte, Brazil and the Universit\'e Paris-Est/Groupe ESIEE, Paris, France.
He is an associate professor at the Computing Department (DECOM), Universidade Federal de Ouro Preto (UFOP), Ouro Preto, Brazil, since August 2008. 
From June 2013 to May 2014 he has been on his sabbatical year at Institute of Computing, University of Campinas (UNICAMP), Campinas, Brazil, as a post-doc / collaborator researcher supported by FAPESP (grant 2013/4172-0).
Currently, he is working as a permanent and collaborator professor at the Post-Graduate Program in Computer Science DECOM-UFOP and DCC-UFMG, respectively.

His research interests include image processing, pattern recognition, computer vision, and information retrieval.
\end{IEEEbiography}

\begin{IEEEbiography}[{\includegraphics[width=1in,height=1.25in,clip,keepaspectratio]{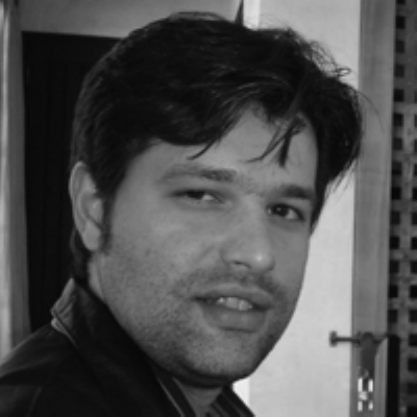}}]{Giovani Chiachia}  
Giovani Chiachia is a post-doctorate researcher at University of Campinas, Brazil.
His main research interest is to approach computer vision and other artificial intelligence tasks with brain-inspired machine learning techniques. 
He received his B.Sc. (2005) and his M.S. (2009) degrees from São Paulo State University and his Ph.D. in Computer Science from the University of Campinas (2013).
He held research scholar positions at the Fraunhofer Institute IIS and at Harvard University as part of his graduate work, and was awarded first place for the performance of his face recognition system in the IEEE Intl. Conf. on Biometrics (2013). 
Beyond his research experience, Dr. Chiachia has a large experience in the IT industry, managing and being part of teams working on projects from a wide range of complexities and scales.
\end{IEEEbiography}

\begin{IEEEbiography}[{\includegraphics[width=1in,height=1.25in,clip,keepaspectratio]{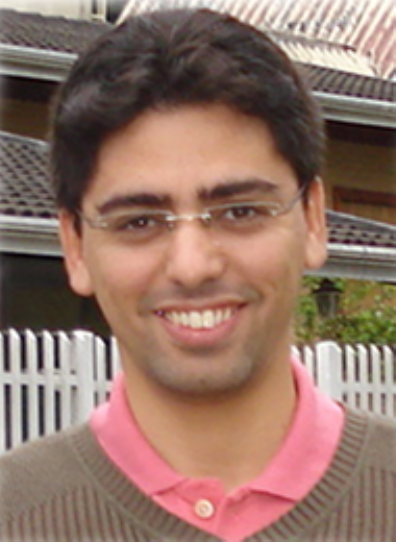}}]{Allan da Silva Pinto} 
Allan Pinto received the B.Sc. degree in Computer Science from University of São Paulo (USP), Brazil, in 2011, and the M.Sc. degree in Computer Science from University of Campinas (Unicamp), Brazil, in 2013. Currently, he is a Ph.D. Student, also in Computer Science, at Institute of Computing, Unicamp, Brazil. 

His research focuses on the areas of image and video analysis, computer forensics, pattern recognition, and computer vision in general, with a particular interest in spoofing detection in biometric systems.
\end{IEEEbiography}

\begin{IEEEbiography}[{\includegraphics[width=1in,height=1.25in,clip,keepaspectratio]{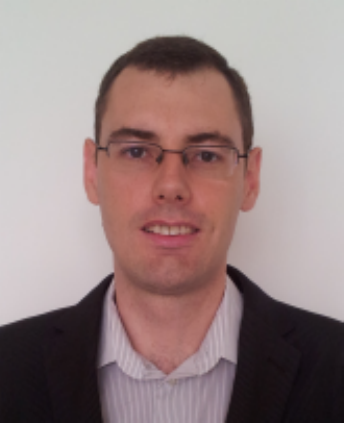}}]{William Robson Schwartz} 
William Robson Schwartz received his Ph.D. degree in Computer Science from University of Maryland, College Park, USA. He received his B.Sc. and M.Sc. degrees in Computer Science from the Federal University of Parana, Curitiba, Brazil.
He is currently a professor in the Department of Computer Science at the Federal University of Minas Gerais, Brazil.

His research interests include computer vision, computer forensics, biometrics and image processing.
\end{IEEEbiography}

\begin{IEEEbiography}[{\includegraphics[width=1in,height=1.25in,clip,keepaspectratio]{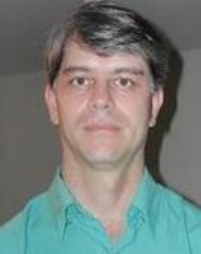}}]{Helio Pedrini} 
Helio Pedrini received his Ph.D. degree in Electrical and Computer Engineering from Rensselaer Polytechnic Institute, Troy, NY, USA. 
He received his M.Sc. degree in Electrical Engineering and his B.Sc. in Computer Science, both from the University of Campinas, Brazil. 
He is currently a professor in the Institute of Computing at the University of Campinas, Brazil. 

His research interests include image processing, computer vision, pattern recognition, machine learning, computer graphics, computational geometry.
\end{IEEEbiography}

\begin{IEEEbiography}[{\includegraphics[width=1in,height=1.25in,clip,keepaspectratio]{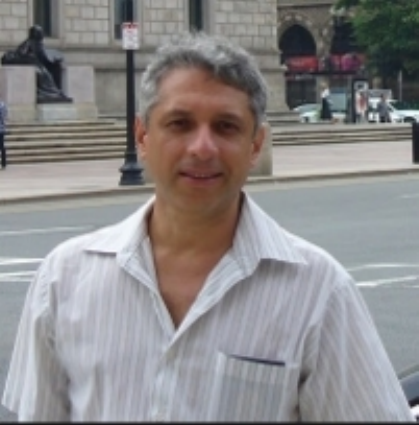}}]{Alexandre Xavier Falcão} 
Alexandre Xavier Falcao is full professor at the Institute of Computing, University of Campinas, Campinas, SP, Brazil. 
He received a B.Sc. in Electrical Engineering from the Federal University of Pernambuco, Recife, PE, Brazil, in 1988. He has worked in biomedical image processing, visualization and analysis since 1991. 
In 1993, he received a M.Sc. in Electrical Engineering from the University of Campinas, Campinas, SP, Brazil. 
During 1994-1996, he worked with the Medical Image Processing Group at the Department of Radiology, University of Pennsylvania, PA, USA, on interactive image segmentation for his doctorate. 
He got his doctorate in Electrical Engineering from the University of Campinas in 1996. 
In 1997, he worked in a project for Globo TV at a research center, CPqD-TELEBRAS in Campinas, developing methods for video quality assessment. 
His experience as professor of Computer Science and Engineering started in 1998 at the University of Campinas. 

His main research interests include image/video processing, visualization, and analysis; graph algorithms and dynamic programming; image annotation, organization, and retrieval; machine learning and pattern recognition; and image analysis applications in Biology, Medicine, Biometrics, Geology, and Agriculture.
\end{IEEEbiography}

\begin{IEEEbiography}[{\includegraphics[width=1in,height=1.25in,clip,keepaspectratio]{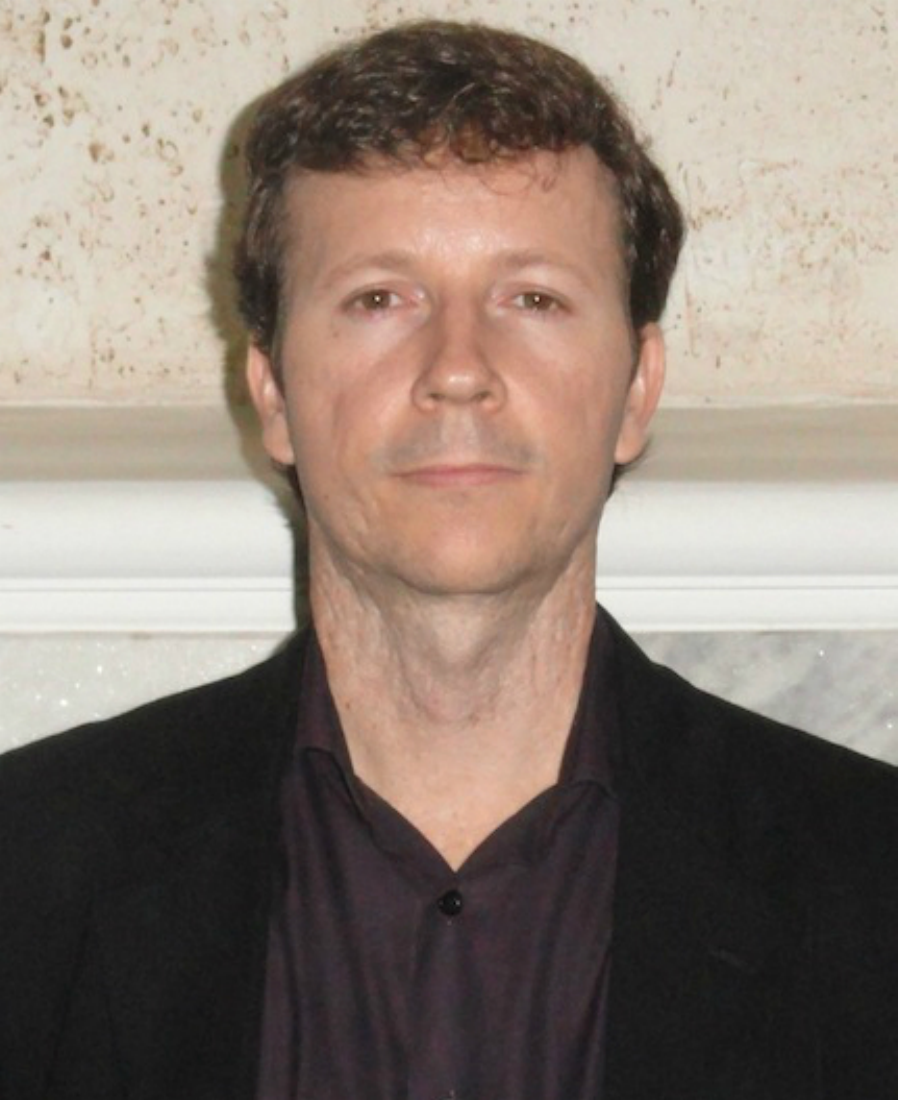}}]{Alexandre de Rezende Rocha}
Anderson de Rezende Rocha received his B.Sc (Computer Science) degree from Federal University of Lavras (UFLA), Brazil in 2003. He received his M.S. and Ph.D. (Computer Science) from University of Campinas (Unicamp), Brazil, in 2006 and 2009, respectively. 
Currently, he is an assistant professor in the Institute of Computing, Unicamp, Brazil. 

His main interests include digital forensics, reasoning for complex data, and machine  intelligence. 
He has actively worked as a program committee member in several important Computer Vision, Pattern Recognition, and Digital Forensics events. 
He is an associate editor of the Elsevier Journal of Visual Communication and Image Representation (JVCI) and a Leading Guest Editor for EURASIP/Springer Journal on Image and Video Processing 
(JIVP) and IEEE Transactions on Information Forensics and Security (T.IFS). 
He is an affiliate member of the Brazilian Academy of Sciences (ABC) and the Brazilian Academy of Forensics Sciences (ABCF). 
He is also a Microsoft Research Faculty Fellow and a former member of the IEEE Information Forensics and Security Technical Committee (IFS-TC).
\end{IEEEbiography}

\end{document}